\documentclass[journal=jacsat,manuscript=article]{achemso}
\usepackage{amssymb}

\usepackage[version=3]{mhchem} 
\usepackage{algorithm}
\usepackage{algpseudocode}

\author{Yayati Jadhav}
\affiliation[CMU]
{Department of Mechanical Engineering, Carnegie Mellon University, Pittsburgh, PA, USA}

\author{Amir Barati Farimani*}
\email{barati@cmu.edu}
\affiliation{
  Department of Mechanical Engineering, Carnegie Mellon University, Pittsburgh,
  PA, USA
}
\alsoaffiliation{
  Machine Learning Department, Carnegie Mellon University, Pittsburgh, PA, USA
}

\title[An \textsf{achemso} demo]
  {LinkD: AutoRegressive Diffusion Model for Mechanical Linkage Synthesis}

\abbreviations{IR,NMR,UV}

\begin{document}







\begin{abstract}

Designing mechanical linkages to produce a target end-effector trajectory remains a fundamental challenge due to the tight interplay between continuous node placements, discrete topological choices, and nonlinear kinematic constraints. The highly nonlinear motion-to-configuration relationship means small perturbations in joint positions can drastically alter trajectories, while the combinatorially expanding design space with increasing node count renders traditional optimization and heuristic methods computationally intractable.
We introduce an autoregressive diffusion framework that leverages the dyadic nature of linkage assembly by representing each mechanism as a sequentially constructed graph where nodes correspond to joints and edges to rigid links. The framework combines a causal transformer with a Denoising Diffusion Probabilistic Model (DDPM), both conditioned on the target end-effector trajectory encoded via a transformer encoder. The causal transformer autoregressively predicts discrete topology node by node, while the DDPM refines each generated node to determine its spatial position and edge connections to previously generated nodes. This sequential generation strategy enables a trial-and-error approach where problematic nodes exhibiting locking behavior or internal collisions can be regenerated, allowing the framework to self-correct degenerate configurations during synthesis.
This graph-based, data-driven approach surpasses traditional optimization and heuristic methods, enabling scalable and efficient inverse design of mechanical linkages that generalize to mechanisms with arbitrary node counts. We demonstrate this capability by successfully synthesizing upto 20 node linkage systems but can be extended to N-nodes. This work advances both autoregressive graph generation methodologies and kinematic synthesis, opening new directions for scalable inverse design of complex mechanical systems.

\end{abstract}

\section{Introduction}


Mechanical linkages have served as the fundamental building blocks of engineered systems for centuries, enabling the transformation of basic rotary input into coordinated, multi-joint motion, providing a reliable foundation for precise and efficient control \cite{reuleaux2013kinematics}. Today, planar linkage mechanisms remain ubiquitous across diverse applications, including robotic manipulators \cite{xin2014control, gallardo2022mechanisms}, automated manufacturing and assembly operations \cite{benhabib1991mechanical, kozuka2013compliant,zhu2015novel}, automotive systems \cite{wu2017workspace}, aerospace mechanisms \cite{ma2022recent}, biomedical devices and prosthetics \cite{lovasz2014experimental, tran2022lightweight}, among others \cite{muller1996novel,zhao2016planar,phocas2020kinematics}. The scope and sophistication of these applications have expanded significantly due to advances in computational design and optimization, which have pushed linkage synthesis well beyond traditional four-bar configurations to enable complex architectures such as six-bar and eight-bar linkages that minimize actuator requirements, reduce system mass and cost, and enhance dynamic behavior \cite{lipson2008evolutionary,ramezani2016bat,plecnik2017design}.

Despite these advances, kinematic synthesis remains computationally challenging due to the simultaneous optimization of discrete topology and continuous geometry. The synthesis problem decomposes into two coupled tasks: selecting the linkage topology (number and types of nodes and their connectivity) and determining geometric parameters (joint positions and link dimensions) \cite{plecnik2014numerical, kim2014topology}. The combinatorial space grows exponentially, with approximately 1.5 million valid six-bar configurations \cite{plecnik2014numerical} and over 4 million mechanisms with up to six loops \cite{tuttle1996generation}. The highly nonlinear mapping between geometry and motion further complicates synthesis, as small dimensional changes can dramatically alter trajectories \cite{vasiliu2001dimensional,lee2024deep}. Additionally, synthesized mechanisms frequently exhibit motion defects such as circuit and branch defects that prevent continuous operation through desired task positions \cite{balli2002defects}.

Traditional approaches to linkage synthesis have addressed these challenges through analytical and optimization-based methods, each with significant limitations. Analytical methods rooted in kinematic theory, such as Burmester theory and Freudenstein's equations \cite{mccarthy2000geometric}, provide closed-form solutions for simple four-bar path synthesis through prescribed precision points, while homotopy continuation methods extend this capability to nine precision points for four-bar linkages \cite{wampler1992complete} and eleven points for six-bar linkages \cite{plecnik2017design}. However, these analytical approaches are restricted to a small number of precision points and cannot handle higher-order mechanisms due to severe computational complexity. Numerical optimization techniques, including mixed integer conic programming (MICP) \cite{pan2023joint}, genetic algorithms \cite{lipson2008evolutionary}, simulated annealing, and evolutionary strategies \cite{ramezani2016bat}, can theoretically handle more complex linkages but are computationally expensive (often requiring hours for a single solution), exhibit strong dependence on initial conditions leading to local minima entrapment, and critically, spend the majority of their computational effort exploring constraint-violating designs that fail to satisfy the strict geometric and kinematic requirements for valid linkage mechanisms \cite{fogelson2023gcp}. Most fundamentally, these approaches require the linkage topology to be specified a priori, forcing designers to manually enumerate and optimize each candidate topology separately. This separation of topology selection and geometric optimization prevents efficient exploration of the coupled design space and limits the discovery of optimal solutions that may require non-intuitive topological structures.

Recent advances in machine learning and deep learning have emerged as promising alternatives to address the computational limitations of traditional approaches. Early neural network-based methods utilized feedforward architectures trained on mechanism datasets to learn mappings between trajectory representations (such as Fourier descriptors or Cartesian coordinates) and mechanism parameters \cite{vasiliu2001dimensional}. These approaches demonstrated near-instantaneous synthesis once trained, offering significant speed advantages over iterative optimization methods. More sophisticated deep neural network architectures have been developed to improve synthesis accuracy, including multi-network ensemble approaches that partition the solution space to handle the inherent multi-modality of the design problem and convolutional neural networks that process image-based trajectory representations \cite{deshpande2021image}. Graph neural networks have been employed to learn representations of mechanism topologies and their kinematic properties \cite{nobari2024link}. Building on this, transformer-based sequence models frame mechanism synthesis as a conditional generation task, processing trajectory information through attention mechanisms to generate both topology and geometric parameters \cite{bolanos2025mechaformer}. Despite these advances, current supervised learning approaches face critical limitations. Most methods are restricted to predefined mechanism types (typically four-bar or six-bar linkages) and require separate models for different topological configurations \cite{deshpande2019computational,khan2015dimensional}, limiting their ability to explore the broader design space. Furthermore, these approaches typically produce single predictions rather than diverse solution sets, failing to capture the one-to-many nature of the synthesis problem where multiple distinct mechanisms can satisfy the same trajectory requirement. The reliance on large labeled datasets also constrains their applicability to novel mechanism types not represented in training data.

Generative models, particularly variational autoencoders (VAEs) and generative adversarial networks (GANs), have been explored to address the limitations of discriminative supervised learning approaches by learning to model the underlying distribution of mechanism designs. Variational autoencoders consist of an encoder network that maps input mechanisms to a lower-dimensional latent space and a decoder network that reconstructs mechanisms from latent representations, enabling the generation of novel designs by sampling from the learned latent distribution \cite{deshpande2019computational}. Conditional VAEs extend this framework by incorporating trajectory information as conditioning input, allowing the model to generate mechanisms that approximate specified paths while exploring variations in the latent space \cite{deshpande2021image}. Generative adversarial networks employ a competitive training process between a generator network that creates synthetic mechanism designs and a discriminator network that distinguishes between real and generated samples, with both networks improving iteratively until the generator produces realistic mechanisms \cite{goodfellow2014generative,lee2024deep}. However, these generative approaches exhibit significant shortcomings for mechanism synthesis applications. VAE-based methods often produce designs with limited diversity and suffer from reconstruction errors that result in kinematically infeasible mechanisms, with reported success rates indicating that less than half of generated designs achieve satisfactory trajectory matching \cite{nurizada2025path}. GANs face training instability issues and mode collapse, where the generator fails to capture the full diversity of the mechanism design space and instead produces repetitive solutions. More critically, both VAE and GAN approaches struggle to maintain kinematic validity throughout the generation process, frequently producing mechanisms that violate geometric constraints or exhibit motion defects such as branch points, dead positions, or mechanical interference. These models also typically treat topology selection and dimensional optimization as coupled but implicit processes, lacking explicit mechanisms to ensure generated designs satisfy the strict requirements for functional single-degree-of-freedom operation.

Reinforcement learning approaches have been explored for linkage synthesis, framing the problem as sequential assembly through graph grammar operations. Lipson\cite{lipson2008evolutionary} pioneered the use of compositional grammar rules with genetic programming to synthesize compound two-dimensional mechanisms, demonstrating human-competitive solutions on classical kinematic challenges such as the straight-line mechanism problem. Building on this foundation, Vermeer et al.\cite{vermeer2018kinematic} employed RL to learn policies for applying these grammar rules, yet their method underperformed compared to Lipson's original evolutionary approach and failed to guarantee kinematically valid designs throughout the entire trajectory. More recent RL-based frameworks like GCP-HOLO\cite{fogelson2023gcp} formulate linkage synthesis as a sequential decision-making problem, using a PPO-trained agent\cite{schulman2017proximal} with graph neural networks to perform tree search over grammar rules that maintain valid 1DOF mechanisms at each assembly step. Similar RL strategies have been applied to co-optimize morphology and control in modular robotic systems\cite{whitman2020modular}, demonstrating the potential for coupled design space exploration in reconfigurable mechanisms. Despite these advances, RL methods face fundamental limitations that constrain their practical applicability. First, they produce solutions that represent local minima in the design space, typically requiring post-processing with global optimization techniques such as CMA-ES\cite{hansen2001completely} to refine geometric parameters and escape suboptimal regions. Second, the sequential nature of RL results in severe sample inefficiency, often demanding hundreds of thousands to millions of design evaluations during training, with computational costs ranging from hours to days for a single synthesis task. Third, RL agents exhibit limited generalization beyond their training distribution, necessitating complete retraining when target paths deviate significantly from those encountered during training. Fourth, the sparse and delayed reward signals inherent to mechanism synthesis complicate credit assignment, frequently leading to premature convergence on suboptimal topologies. Finally, designing reward functions that effectively balance the competing objectives of kinematic feasibility, path accuracy, and mechanical constraints (such as link length ratios, transmission angles, and motion defects) remains a persistent challenge, with minor modifications to reward structure often producing dramatically different convergence behaviors. These compounding limitations motivate the development of direct generative approaches that can jointly model topology and geometry in a single forward pass, eliminating the need for iterative sequential assembly procedures and enabling efficient exploration of the coupled design space.

Graph neural networks (GNNs)\cite{wu2020comprehensive,corso2024graph,scarselli2008graph} can be leveraged to process linkage graphs, learning meaningful representations that capture both local joint connectivity and global motion constraints. By embedding linkage configurations into a latent graph space, a GNN can learn to generate structurally sound designs, ensuring connectivity and functional constraints are preserved \cite{wu2020comprehensive,zhou2020graph,liu2022introduction}. Furthermore, GNNs can generalize to different linkage topologies, allowing for the synthesis of mechanisms beyond simple four-bar or six-bar configurations. However, existing GNN-based approaches typically focus on generating either the nodal positions or the adjacency matrix, but rarely both simultaneously. This limitation highlights an open challenge: developing graph representations and learning frameworks that can jointly capture topology and geometry, enabling the network to learn both structure and spatial configuration in a unified manner. 

To address this challenge, we represent each graph node as a token and leverage the autoregressive capability of causal transformers to model how each node connects to previously generated nodes. This formulation enables the transformer to learn both the sequential dependencies and structural logic of linkage formation. To further refine these outputs, we introduce a DDPM head that iteratively denoises the nodal connection types and spatial positions, transforming coarse predictions into valid and physically consistent configurations.

Denoising Diffusion Probabilistic Models (DDPM) have proven effective across diverse engineering tasks\cite{jadhav2023stressd, li2025latent, jadhav2024generative, zhou2024text2pde, bartsch2024sculptdiff, graves2024airfoil}, from stress field generation and fluid flow prediction to PDE solving and 3D lattice synthesis. This iterative denoising capability naturally suits mechanical design, where constraints like kinematic feasibility and connectivity must be progressively validated and enforced. We leverage this by combining transformers' structural reasoning with diffusion's refinement properties\cite{li2024autoregressive}, creating a unified framework for inverse linkage design. Our approach jointly synthesizes topology and geometry while maintaining generative stability and diversity, grounding the generation process in mechanical principles.

\begin{figure}[!h]
    \includegraphics[width=\textwidth]{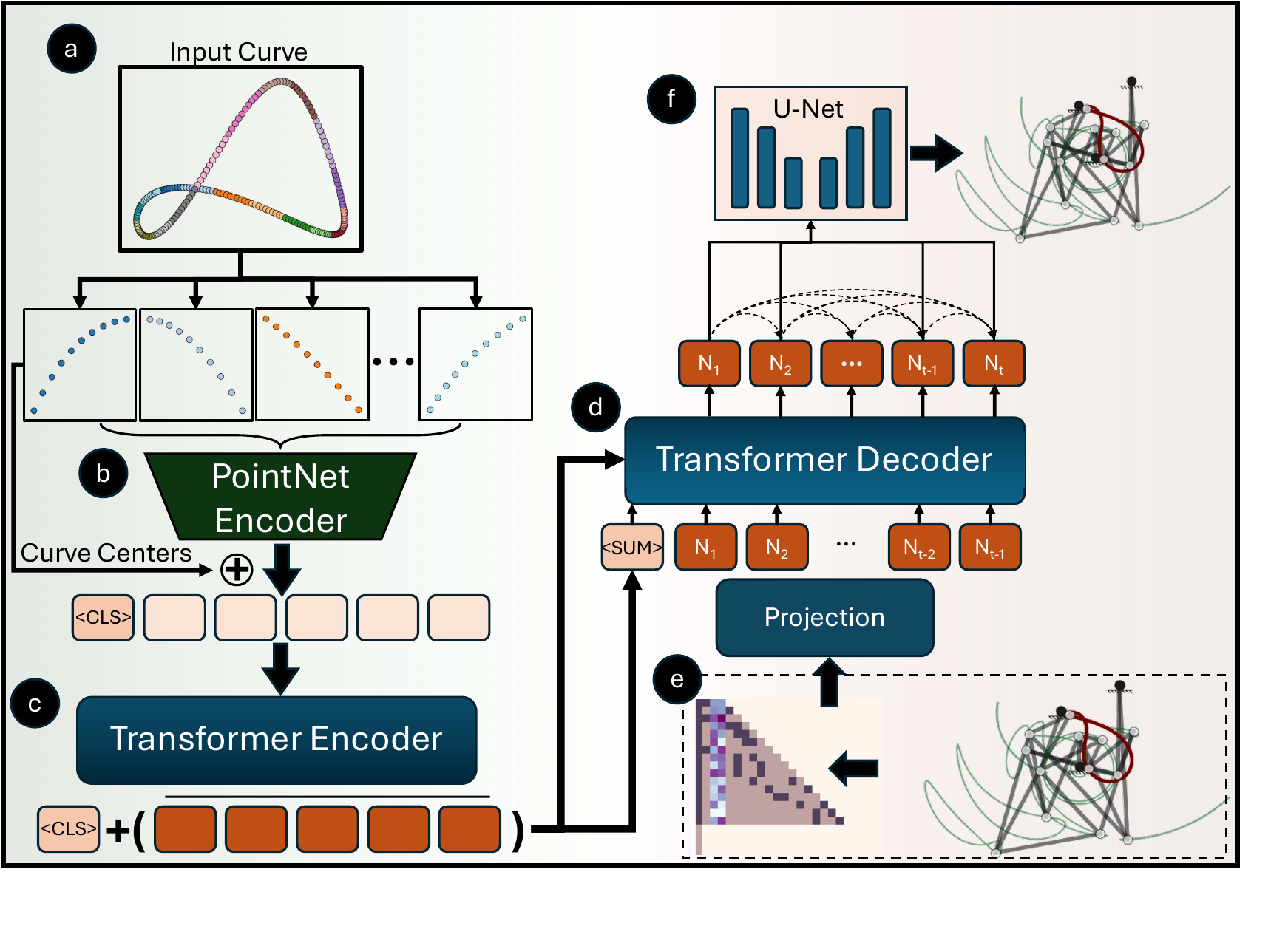}
    \caption{ \textbf{ Overview of the proposed framework.} (a) The input curve is divided into smaller sub-curves, each encoded using (b) a PointNet encoder with curve centers serving as positional encodings. The resulting embeddings are processed by (c) a Transformer encoder with self-attention. The learned “[CLS]” token is combined with the mean sub-curve embedding to form a conditioning vector for an autoregressive causal (d) Transformer decoder, which operates on (e) the graph representation of the mechanism. (f) A DDPM head with a U-Net backbone further refines the Transformer output to regenerate the final graph structure. }
    \label{fig:overview}
\end{figure}

Figure \ref{fig:overview} illustrates the overview of our proposed framework for generating mechanical linkages that reproduce a given end-effector trajectory. The input curve (a) is first divided into segments, and their centers are extracted as point cloud representations. These curve centers are then encoded by a PointNet \cite{qi2017pointnet} encoder (b), which captures the geometric features of the desired trajectory. The encoded features are combined with a learnable \texttt{<CLS>} token and processed by a Transformer encoder \cite{yu2022point} (c) to obtain a contextualized representation of the input trajectory.

The generation process is handled by an autoregressive Transformer decoder (d), which sequentially predicts graph nodes $N_1, N_2, \ldots, N_t$ representing the mechanical linkage structure. Starting from a \texttt{<SUM>} token, the decoder generates each node conditioned on previously generated nodes, thereby learning the sequential dependencies inherent in linkage topology. The decoder outputs include both the connectivity information and initial spatial positions of the nodes (e), forming a graph representation of the mechanical linkage.

To refine these predictions into physically consistent and kinematically valid configurations, a U-Net diffusion model (f) iteratively denoises the graph structure. This refinement process progressively enforces mechanical constraints and ensures that the generated linkage can accurately trace the input trajectory. The final output is a complete mechanical linkage design with defined joint types, link connections, and spatial coordinates that satisfy both topological and geometric requirements.

\section{Methodology}

\subsection{Problem Formulation}

\begin{figure}[!h]
    \includegraphics[width=\textwidth]{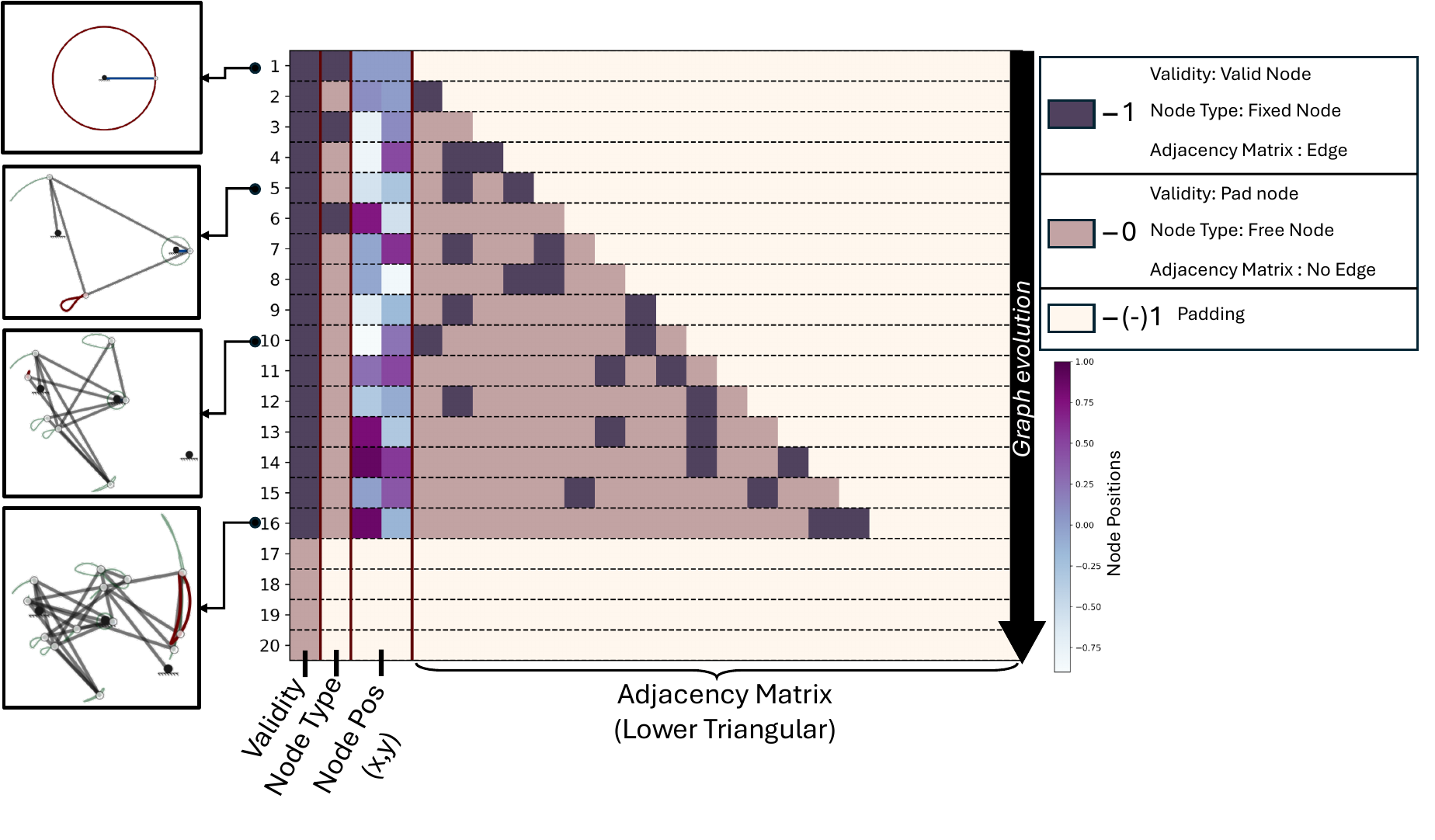}
    \caption{ \textbf{ Graph representation of mechanical linkage.} The mechanism graph is represented as an $N \times 24$ matrix, where
$N$ is the number of nodes. The first entry in each 24-dimensional feature vector encodes node validity, distinguishing valid nodes from padding. The second dimension specifies the node type, followed by the normalized nodal positions in the $x$ and $y$ directions. The remaining entries correspond to the lower-triangular half of the adjacency matrix, representing pairwise node connections. Each subgraph derived from this representation constitutes a valid mechanism instance. }
\label{fig:graph_representation}
\end{figure}

The mechanical linkage synthesis problem for a planar one degree of freedom (1-DoF) mechanism is formulated as follows: given a target coupler curve $\mathcal{C} \subset \mathbb{R}^2$, the objective is to generate a feasible planar linkage $G$ such that the simulated motion reproduces $\mathcal{C}$ within a specified tolerance. 

To enable learning-based synthesis, each mechanism is represented as a directed graph $G = (V, X, E, T)$, where the components encode both topological and geometric properties required to define a valid linkage. As illustrated in Figure~\ref{fig:graph_representation}, this representation accommodates mechanisms of varying complexity within a unified framework by organizing graph components into a structured $N \times 24$ feature matrix, where $N$ denotes the maximum number of nodes.

The node validity vector $V \in \{0,1\}^n$ distinguishes active nodes ($V_i = 1$) from padded or inactive nodes ($V_i = 0$), enabling variable-size mechanisms within a fixed-dimension representation. The type vector $T \in \{0,1,-1\}^n$ specifies discrete joint classifications, where $T_i = 1$ indicates a grounded (fixed) joint, $T_i = 0$ denotes a revolute joint, and $T_i = -1$ represents a padded position. The positional matrix $X \in [-1,1]^{n \times 2}$ specifies the continuous spatial coordinates of each joint in normalized planar space. 

Critically, the adjacency matrix $E \in \{0,1,-1\}^{n \times n}$ is represented using only its lower triangular entries, as the connectivity between joints is symmetric. This design choice reduces redundancy while preserving complete topological information. Furthermore, this lower triangular representation naturally induces a causal structure during generation, where each node's connectivity depends solely on previously generated nodes, enabling sequential graph construction through autoregressive modeling. The 24-dimensional feature vector for each node comprises the validity flag, node type, spatial coordinates ($x$, $y$), and the corresponding row from the lower triangular adjacency matrix, creating a compact yet complete specification.

This graph-based formulation provides a unified framework that captures the complete kinematic specification of a planar linkage. A feasible linkage must satisfy both kinematic and geometric constraints to ensure physically valid motion, as detailed in Section~\ref{sec:val_cri}.
\subsection{Auto-regressive Diffusion}

A distinctive property of planar mechanical linkages is that they are dyadically compositional. Any subgraph of a valid linkage remains a valid mechanism because linkages are constructed from dyads, which are two-link, three-joint building blocks that can be recursively combined while preserving kinematic validity. For instance, a complete 20-node mechanism can be decomposed into smaller sub-linkages containing 2, 3, \dots, 19 nodes, each satisfying the Gruebler–Kutzbach mobility criterion. This hierarchical composition property distinguishes linkage graphs from other design domains, such as molecular~\cite{yamada2023molecular,liu2021graphebm} or circuit graphs~\cite{dong2023cktgnn}, where intermediate subgraphs often lack functional meaning. In contrast, every partial linkage corresponds to a physically interpretable sub-mechanism, making the problem naturally suited to an autoregressive generation paradigm where the model constructs mechanisms node by node while preserving mechanical feasibility at each step.

Generating a complete linkage in a single step often leads to kinematically invalid or overconstrained configurations. Small geometric errors in predicted node coordinates can propagate through the graph, violating geometric consistency (e.g., link-length or joint constraints) and resulting in immobile or locked structures. To address this, our framework adopts a causal Transformer backbone that generates node embeddings in sequence, ensuring that each prediction depends only on previously generated nodes. At generation step $t$, the Transformer attends to all nodes $< t$ using a causal mask, thereby maintaining strict autoregressive order and embedding the structural and kinematic history of the partial mechanism.

From the causally generated node-wise graph embeddings, a conditional DDPM~\cite{ho2020denoising,song2020denoising} equipped with a U-Net–style MLP architecture iteratively refines and projects these latent representations into the original graph space. Given that each node follows a fixed structural template, the model is supervised with complementary loss functions tailored to each attribute. A weighted cross-entropy loss is applied for adjacency prediction to determine edge existence, while additional classification losses handle node validity and type. For continuous geometry, a Huber loss supervises the predicted node positions, providing robustness to outliers during spatial refinement. This joint formulation enables the DDPM to output all node attributes in a single unified diffusion process. The model generates both discrete structural components (node type $T_t$, validity $V_t$, and adjacency entries $E_{t,1:t-1}$) and continuous spatial coordinates $X_t = (x_t, y_t)$, reconciling discrete relational structure and continuous spatial configuration within a flexible training framework.

\subsubsection{Inference Procedure}
During inference, linkage generation proceeds sequentially under the autoregressive diffusion framework~\cite{li2024autoregressive}. Given a target coupler curve $\mathcal{C}$, the model constructs a valid linkage graph $G = (V, X, E, T)$ by generating one node at a time, together with its connections to previously generated nodes. Let $\mathbf{T}_t$ denote the feature vector of the $t$-th node, containing its position $(x_t, y_t)$, type $T_t$, validity $V_t$, and the lower-triangular adjacency entries $E_{t,1:t-1}$ that define its links to nodes $1, \dots, t-1$. The generative process factorizes autoregressively as:
\begin{equation}
    p(G \mid \mathcal{C}) 
    = \prod_{t=1}^{N} p\!\left( \mathbf{T}_t \,\middle|\, \mathbf{T}_{<t}, \mathcal{C} \right),
    \label{eq:autoregressive_factorization}
\end{equation}
where $N$ is the maximum number of nodes (set to 20) and $\mathbf{T}_{<t}$ represents the partial mechanism generated up to step $t-1$.

At each generation step $t$, the causal Transformer encodes the current partial graph $\mathbf{T}_{<t}$ together with the target trajectory embedding $E(\mathcal{C})$ to produce a contextualized node embedding. The conditional DDPM then samples all node attributes jointly from this embedding. Once the partial graph contains more than two nodes, each proposed node $\mathbf{T}_t$ undergoes kinematic validation through forward simulation, which verifies the 1-DoF constraint and checks for geometric degeneracies such as overlapping joints or invalid link configurations. If validation succeeds, the node is appended to the partial graph and generation proceeds to the next step. If validation fails, a node-level retry mechanism is invoked to resample the configuration, as detailed in the following subsection.

\subsubsection{Node-Level Retry Mechanism}
To handle prediction errors while maintaining generation efficiency, we implement a bounded retry strategy that selectively resamples invalid nodes without restarting the entire generation process. When kinematic validation fails for a newly proposed node, the model resamples only that node using the same conditional context, leaving all previously validated nodes unchanged. This approach is particularly effective when generating batches of mechanisms simultaneously, as it enables targeted correction of individual failures within the batch.

The retry mechanism is constrained to a maximum of $K=10$ attempts per node. If a valid configuration cannot be obtained within this limit, generation proceeds with the best available sample. This design prevents the generation process from stalling on particularly challenging constraint configurations while still providing multiple opportunities to recover from local prediction errors. Algorithm~\ref{alg:node_retry} formalizes this procedure.

\begin{algorithm}[t]
\caption{Node-Level Retry with Kinematic Validation}
\label{alg:node_retry}
\begin{algorithmic}[1]
\Require Target curve $\mathcal{C}$, maximum nodes $N_{\max}=20$, max retries $K=25$
\Ensure Generated linkage graph $G$
\State Initialize partial graph $\mathbf{T}_{<1} \gets \emptyset$
\State Encode target curve: $\mathbf{c} \gets E(\mathcal{C})$
\For{$t = 1$ to $N_{\max}$}
    \State $\mathbf{h}_t \gets \textsc{CausalTransformer}(\mathbf{T}_{<t}, \mathbf{c})$ \Comment{Encode context}
    \State $\mathbf{T}_t \gets \textsc{DDPM-Sample}(\mathbf{h}_t)$ \Comment{Sample node attributes}
    \If{$t > 2$}
        \State $\mathit{valid} \gets \textsc{KinematicValidation}(\mathbf{T}_{\leq t})$
        \State $\mathit{attempt} \gets 0$
        \While{$\mathit{not valid}$ \textbf{and} $\mathit{attempt} < K$}
            \State $\mathbf{T}_t \gets \textsc{DDPM-Sample}(\mathbf{h}_t)$ \Comment{Resample only node $t$}
            \State $\mathit{valid} \gets \textsc{KinematicValidation}(\mathbf{T}_{\leq t})$
            \State $\mathit{attempt} \gets \mathit{attempt} + 1$
        \EndWhile
        \If{$\neg \mathit{not valid}$}
            \State \textbf{Warning:} Node $t$ remains invalid after $K$ retries
        \EndIf
    \EndIf
    \State $\mathbf{T}_{<t+1} \gets \mathbf{T}_{<t} \cup \{\mathbf{T}_t\}$ \Comment{Append to partial graph}
\EndFor
\State \Return $G = (\mathbf{V}, \mathbf{X}, \mathbf{E}, \mathbf{T})$ from $\mathbf{T}_{\leq N_{\max}}$
\end{algorithmic}
\end{algorithm}

For batch generation, the algorithm extends naturally to handle multiple mechanisms simultaneously. When validation fails for a subset of samples in the batch, only those indices are resampled while valid samples remain fixed. This selective resampling strategy significantly improves computational efficiency compared to full-batch regeneration, as most samples typically require few or no retries. The bounded retry limit ensures that generation completes within a predictable time while the iterative validation maintains kinematic feasibility throughout the sequential construction process.

\subsection{Denoising diffusion probabilistic model (DDPM)}\label{sec:ddpm}

Denoising Diffusion Probabilistic Models (DDPM)\cite{ho2020denoising} involve the formulation of a Markov chain of diffusion steps, progressively introducing tractable noise to the input data during the forward process. Given implicit fields as input $x_0$, the forward diffusion process is represented as a Markov chain\cite{ho2020denoising}.

\begin{equation}
\begin{array}{l}
     q(x_t|x_{t-1})=\mathcal{N}(x_t;\sqrt{1-\beta_t}x_{t-1},\beta_t I ) \\
     q(x_{1:T}|x_0)=\prod_{t=1}^T q(x_t|x_{t-1})
\end{array}
\end{equation}

Where $\beta_t\in (0,1)$ is the variance scheduler which controls the step size of noise added.

The goal of the neural network $p$, is to generate meaningful nodal features from Gaussian noise input $x_T \approx \mathcal{N}(0,I)$ given a conditioning vector '$z$'. The generation process (backward process) is the reverse of the forward process, where the neural network $p$, learns to recover the original implicit field $x_{0}$ given $x_t$ and $z$. The backward process can be written as:

\begin{equation}
    p(x_0:T)=p(x_T)\prod^{T}_{t=1} p(x_{0}|x_t, z)
\end{equation}

The diffusion component of our model operates on the node feature vectors $\mathbf{T}_t$ defined in the sequential graph representation.  
At each generation step $t$, the partial graph $\mathbf{T}_{<t}$ and target path embedding $E(\mathcal{C})$ are used as conditioning inputs to a denoising diffusion probabilistic model (DDPM).

\subsection{Network Architecture}

The proposed autoregressive diffusion framework consists of a modified causal Transformer backbone and two task-specific output heads: a topology head for discrete graph structure prediction and a spatial head for continuous nodal position refinement.

\begin{figure}[!h]
    \includegraphics[width=\textwidth]{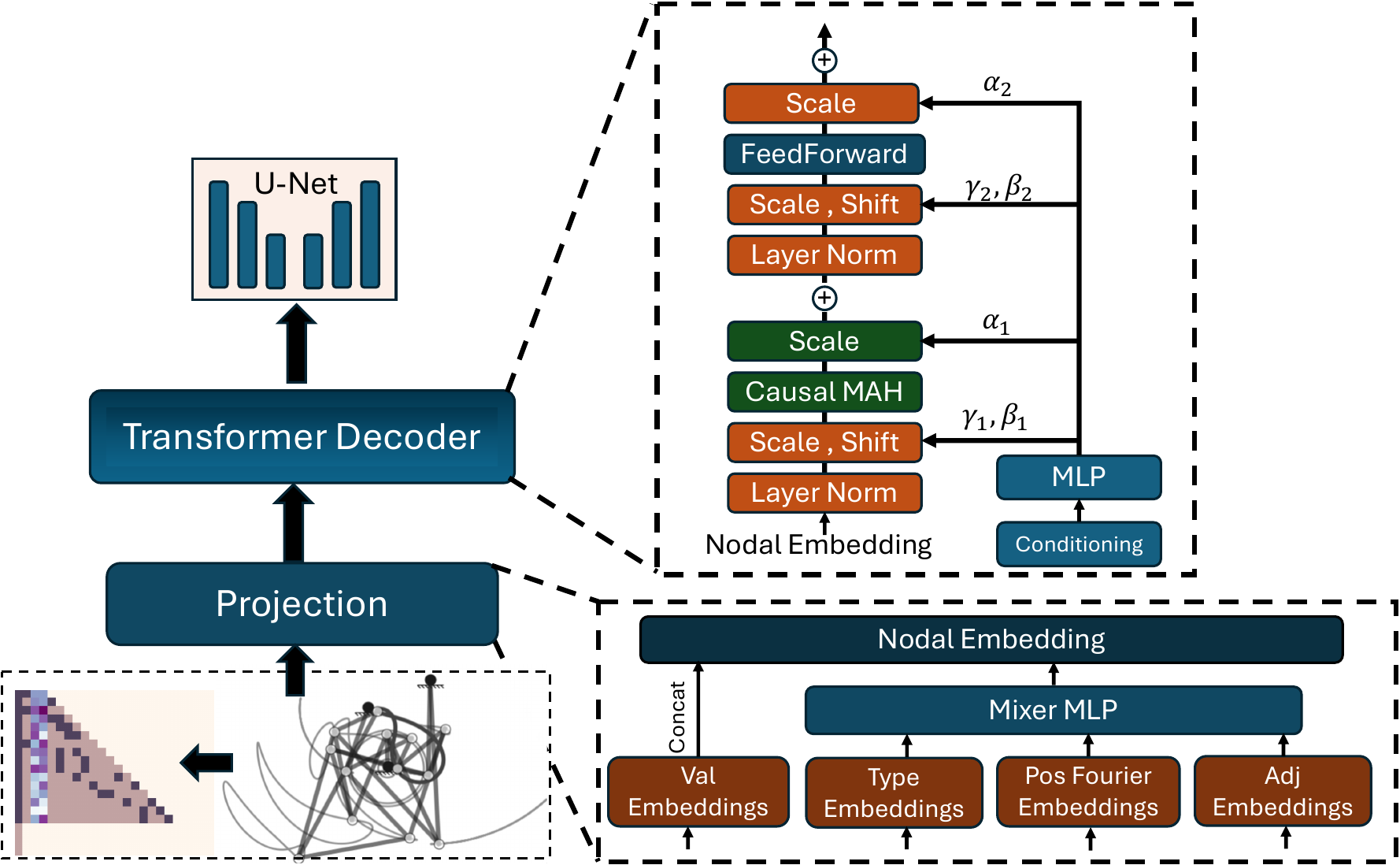}
    \label{figure:peridicity}
    \caption{ \textbf{ Transformer backbone.} The transformer backbone encodes each node using categorical validity embeddings, discrete type embeddings, continuous learned Fourier positional embeddings, and discrete adjacency embeddings. These features are fused through a mixer MLP and concatenated with the validity embeddings to form nodal representations. A diffusion-inspired transformer (DiT) structure is then applied, where each block performs layer normalization, FiLM-style scale–shift modulation, and causal multi-head attention followed by a feed-forward network. The entire model is conditioned on the target curve trajectory, which modulates the transformer layers to guide node generation toward matching the desired motion pattern. }
\end{figure}

\paragraph{Causal Transformer Backbone.}
The Transformer serves as the core sequence model responsible for capturing the evolving structural dependencies of the linkage graph. 
At generation step $t$, the Transformer receives the partial sequence of node features $\mathbf{T}_{<t}$ and produces a contextualized embedding for the next node $\mathbf{T}_t$. 
A causal attention mask is applied so that each node embedding attends only to previously generated nodes, ensuring strict autoregressive ordering. 

To incorporate information about the desired motion, the Transformer is conditioned on the target coupler curve $\mathcal{C}$ through Feature-wise Linear Modulation (FiLM) \cite{perez2018film} layers. 
Given the curve embedding $E(\mathcal{C})$, FiLM applies learned affine transformations to the intermediate activations of the Transformer blocks:
\[
\mathrm{FiLM}(h) = \gamma(E(\mathcal{C})) \odot h + \beta(E(\mathcal{C})),
\]
where $\gamma(\cdot)$ and $\beta(\cdot)$ are small MLPs that scale and shift the hidden representation $h$ at each layer. 
This modulation provides global conditioning while preserving autoregressive consistency, allowing the Transformer to adapt its structural predictions to different target motion trajectories without requiring multiple conditioning tokens.

\section{Results}

\begin{figure}[t]
    \centering
    \includegraphics[width=\linewidth]{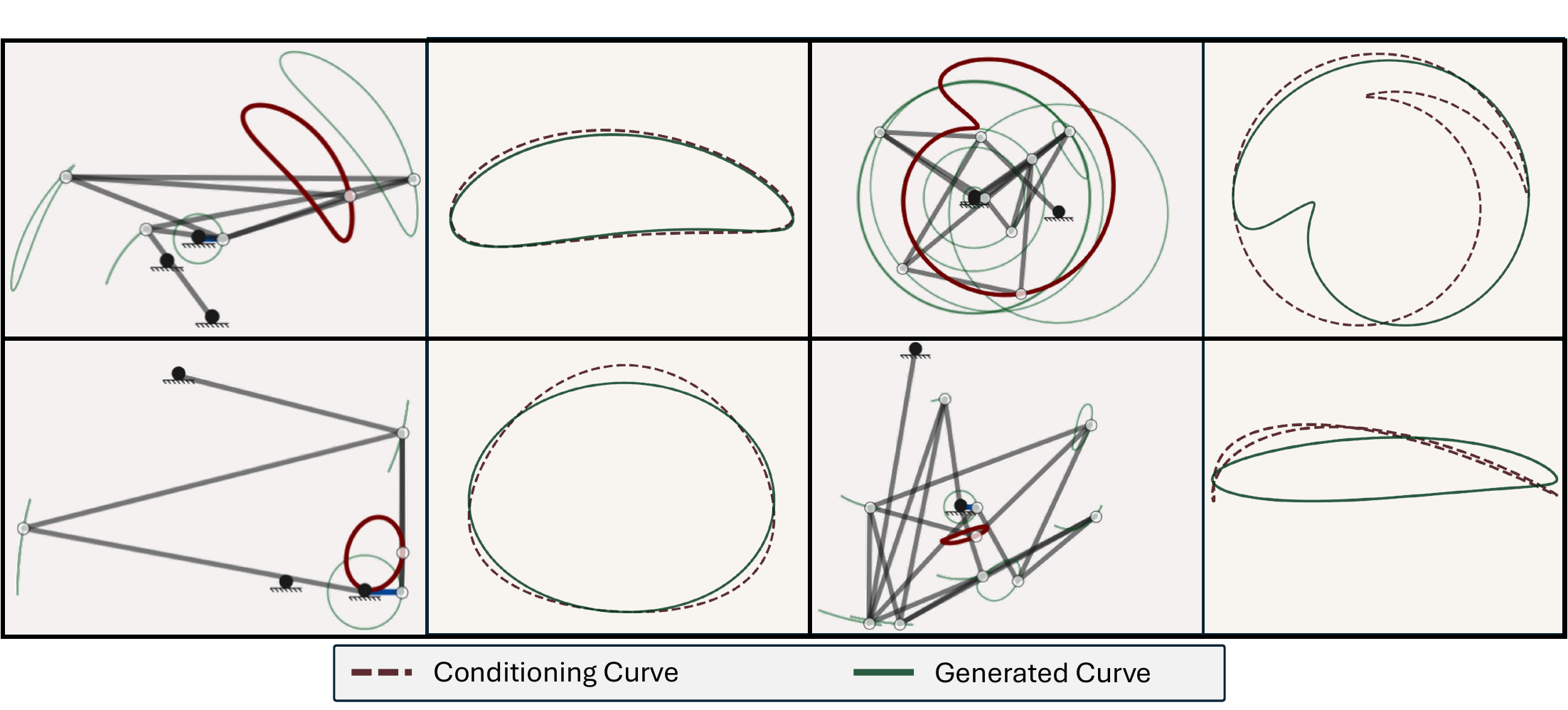}
    \caption{
    \textbf{Qualitative results of mechanism generation conditioned on target curves.} 
    Each row shows a different example with the complete mechanism visualization (left) and curve comparison (right). Dashed lines represent the conditioning curves, while solid green lines show the curves traced by the generated mechanism's end effector. The model successfully generates mechanically valid linkages across diverse curve geometries (elongated ellipses, semi-circular arcs, and circular paths) with close but not perfect alignment between conditioning and generated curves, reflecting the stochastic nature of conditional DDPM generation.
    }
    \label{fig:results}
\end{figure}

We evaluate our model's ability to generate mechanically valid linkages conditioned on target motion trajectories. Figure~\ref{fig:results} presents representative examples showing complete mechanism structures (left) and curve comparisons (right). Dashed lines indicate conditioning curves, while solid green lines show curves traced by the generated mechanism's end effector.
Our model successfully synthesizes diverse linkages across varied curve geometries. Generated mechanisms exhibit appropriate topological complexity, from simple four-bar to complex multi-bar configurations, with all examples satisfying topological validity and kinematic feasibility.
Notably, generated curves do not perfectly replicate conditioning curves, this is expected for conditional DDPM generation. The conditioning signal guides but does not strictly constrain the stochastic diffusion process, allowing exploration of valid mechanisms near the target trajectory. This variability reflects the one-to-many inverse problem: multiple valid linkages can produce similar curves. Despite this inherent variation, generated curves closely approximate conditioning targets while maintaining mechanical validity, demonstrating effective learning of the trajectory-to-mechanism mapping.
The mechanism visualizations confirm correct placement of fixed nodes and joints, with linkages achieving continuous motion throughout their range. Diverse structural solutions for similar curves (e.g., circular trajectories) indicate the model learns multiple design strategies rather than memorizing archetypes, demonstrating robust generalization capability.

\subsection{Generation Success Rate Evaluation}
\begin{figure}[!h]
    \includegraphics[width=\textwidth]{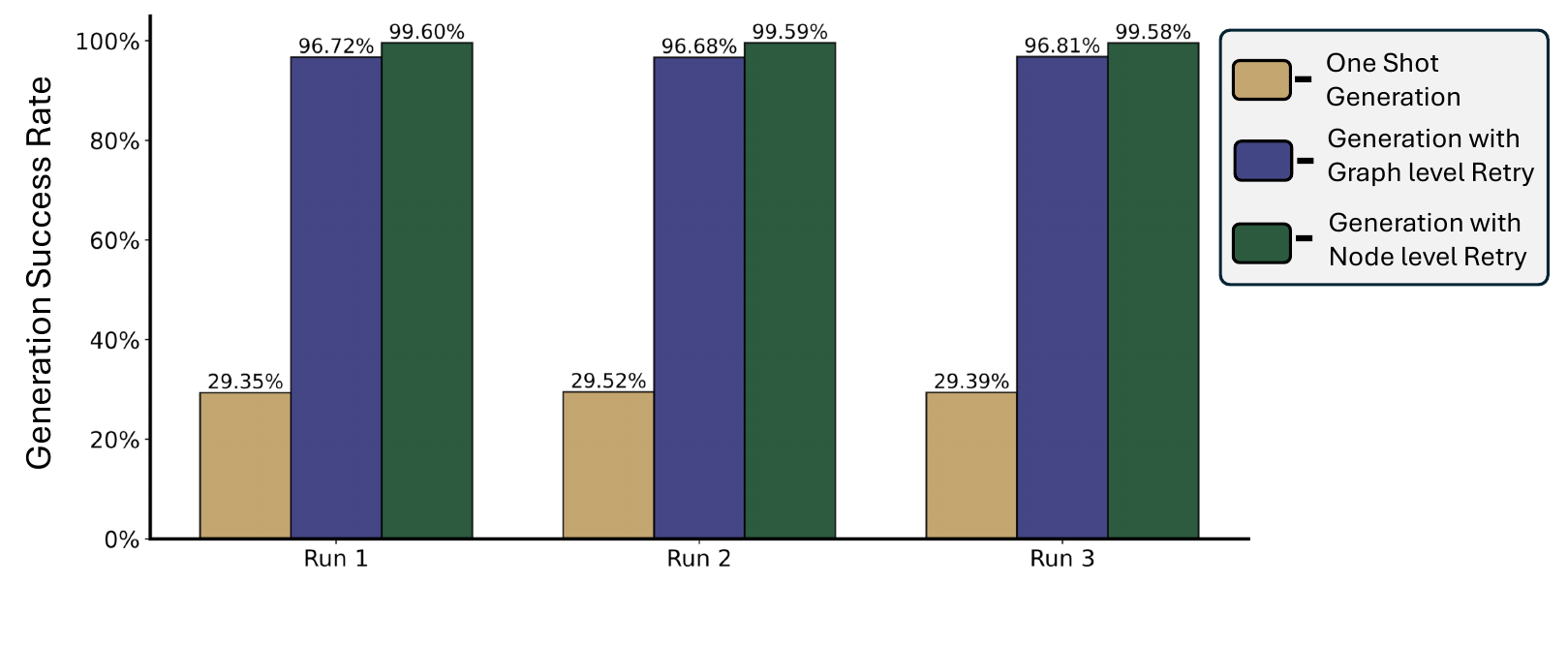}
    
    \caption{ \textbf{ Evaluation of Generation Success Rates Across Retry Strategies.} Comparison of one-shot generation with graph-level and node-level retry mechanisms across three independent experimental runs. Graph-level retry improves success rate from 29.4\% to 96.7\%, while node-level retry achieves 99.6\% success rate. }
    \label{fig:success_rate}
\end{figure}

We systematically evaluate the impact of retry strategies on generation reliability through three experimental conditions. For one-shot generation, we perform a single forward pass through the diffusion model for each test curve without any retry mechanism. For graph-level retry, failed mechanisms are completely regenerated up to 25 times until a valid graph is obtained. For node-level retry, we adopt a more granular approach: each node is generated and immediately validated, with failed nodes regenerated up to 25 times before proceeding to the next node. Since DDPM generation is inherently stochastic, we conduct three independent experimental runs with different random seeds to ensure reproducibility and statistical robustness.

\subsubsection{Validation Criteria}
\label{sec:val_cri}
Generated mechanisms must satisfy two sequential validation checks to be considered successful. First, we verify \textbf{topological validity}, i.e., whether the adjacency matrix $\mathbf{A} \in \{0,1\}^{N \times N}$ represents a valid dyadic mechanism. A mechanism is dyadic if it can be assembled sequentially, where each node's position is uniquely determined by exactly two previously positioned nodes through trilateration. Formally, given a set of fixed nodes $\mathcal{F}$ and motor endpoint $m$, we initialize the known set $\mathcal{K}_0 = \mathcal{F} \cup \{m\}$ and iteratively expand it. At each step $t$, we identify a node $i \notin \mathcal{K}_t$ such that:
\begin{equation}
\left|\{j \in \mathcal{K}_t : A_{ij} = 1\}\right| = 2
\end{equation}
The mechanism passes topological validation if all valid nodes eventually become known ($\mathcal{K}_T = \mathcal{V}$). Failure occurs when: (1) a node has more than 2 known neighbors (overconstrained), or (2) no node with exactly 2 known neighbors exists while unknowns remain (non-dyadic structure).

Second, we verify \textbf{kinematic feasibility}, i.e., whether the mechanism can move through its full range of motion without encountering geometric impossibilities. For each motor angle $\theta \in [0, 2\pi]$, we solve the forward kinematics by iteratively determining node positions. For node $k$ with two known neighbors $i$ and $j$, we apply the law of cosines to find the angle $\phi(\theta)$:
\begin{equation}
\cos\phi(\theta) = \frac{l_{ij}^2(\theta) + G_{ik}^2 - G_{jk}^2}{2 l_{ij}(\theta) G_{ik}}
\end{equation}
where $l_{ij}(\theta) = \|\mathbf{x}_i(\theta) - \mathbf{x}_j(\theta)\|_2$ is the distance between known nodes at angle $\theta$, and $G_{ik}$, $G_{jk}$ are the fixed link lengths. The mechanism is kinematically valid only if $\cos\phi(\theta) \in [-1, 1]$ for all nodes and all 200 sampled angles. Violation of this constraint indicates a \textbf{branch defect}, a configuration where the triangle inequality is violated and the mechanism cannot physically achieve that position. The position is then computed via rotation:
\begin{equation}
\mathbf{x}_k(\theta) = \mathbf{x}_i(\theta) + \mathbf{R}(\phi) \cdot G_{ik} \cdot \frac{\mathbf{x}_j(\theta) - \mathbf{x}_i(\theta)}{l_{ij}(\theta)}
\end{equation}
where $\mathbf{R}(\phi)$ is a 2D rotation matrix. A mechanism passes validation if and only if both topological and kinematic constraints are satisfied.

Figure~\ref{fig:success_rate} presents success rates across all experimental conditions. One-shot generation establishes a baseline success rate of 29.4\% ($\pm$0.09\%), revealing that approximately 70\% of initially generated mechanisms fail either topological or kinematic validation. This relatively low success rate underscores the challenge of satisfying both graph-theoretic constraints (dyadic assembly property) and geometric constraints (branch-free kinematics) simultaneously in a single forward pass.

Graph-level retry demonstrates a dramatic improvement, achieving 96.7\% ($\pm$0.07\%) success rate, a 67.3 percentage point increase over one-shot generation. This substantial gain indicates that most failure cases can be addressed through complete regeneration, suggesting that failures are typically stochastic rather than systematic. The maximum of 25 retry attempts proves sufficient for the vast majority of cases, with most successful generations occurring within the first few attempts.

Node-level retry further refines performance, reaching 99.6\% ($\pm$0.01\%) success rate. The additional 2.9 percentage point improvement represents an 86\% reduction in remaining failures compared to graph-level retry. Critically, node-level retry achieves this superior performance with greater efficiency: by validating each node individually against both topological constraints (connectivity to exactly two known nodes) and kinematic constraints (valid triangle formation) before proceeding, the method preserves valid portions of the generation and only regenerates failed components. This localized correction strategy reduces overall computational overhead compared to discarding and regenerating entire graphs.

The consistency of results across all three independent runs is notable, with maximum standard deviation of 0.09\% across all conditions. This low variance confirms that the observed performance differences between strategies are statistically robust and not artifacts of particular random seeds, demonstrating the reliability of our generative approach for mechanically valid linkage synthesis.

\subsection{Geometric Accuracy of Generated Trajectories}

\begin{figure}[!h]
    \includegraphics[width=\textwidth]{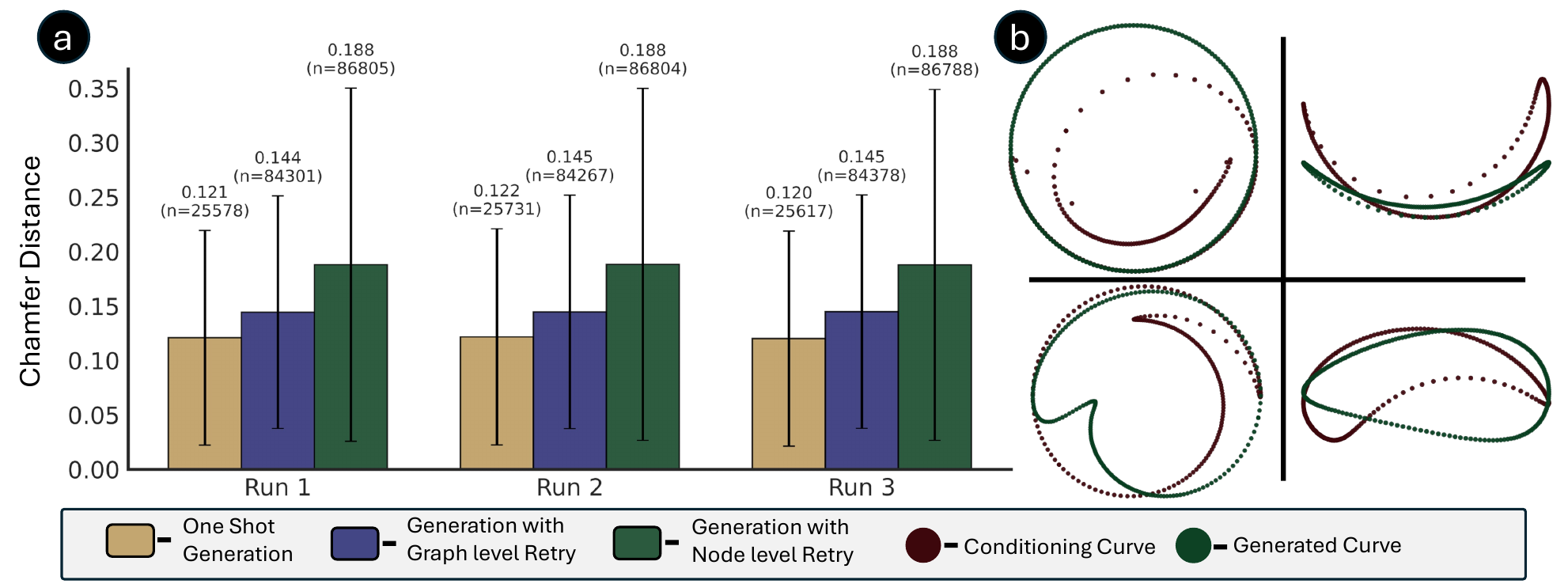}
    
    \caption{\textbf{Chamfer Distance between conditioning and generated curves across retry strategies.} (a) Mean Chamfer Distance with standard deviation error bars computed over three independent runs. Curves are represented by 200 points corresponding to uniformly sampled motor angles. Sample sizes (n) indicate the number of valid mechanisms evaluated for each condition. (b) Representative examples showing conditioning curves (maroon) and generated curves (solid green). Points indicate the actual sampled positions along each curve, demonstrating non-uniform spatial distribution due to mechanism kinematics. Examples also show rotational and reflectional variations that arise even after curve normalization, which contribute to Chamfer Distance measurements.}

    \label{fig:chamfer}
\end{figure}

We evaluate curve reconstruction accuracy using Chamfer Distance measurements between conditioning curves and curves traced by generated mechanisms. Figure~\ref{fig:chamfer}(a) presents results across all three generation strategies. Chamfer Distance quantifies the geometric similarity between point sets, computed as:
\begin{equation}
d_{\text{CD}}(\mathcal{C}_1, \mathcal{C}_2) = \frac{1}{|\mathcal{C}_1|}\sum_{\mathbf{x} \in \mathcal{C}_1} \min_{\mathbf{y} \in \mathcal{C}_2} \|\mathbf{x} - \mathbf{y}\|_2 + \frac{1}{|\mathcal{C}_2|}\sum_{\mathbf{y} \in \mathcal{C}_2} \min_{\mathbf{x} \in \mathcal{C}_1} \|\mathbf{y} - \mathbf{x}\|_2
\end{equation}
where $\mathcal{C}_1$ and $\mathcal{C}_2$ represent the conditioning and generated curve point sets, respectively. Each curve is represented by 200 points obtained by simulating the mechanism through uniformly sampled motor angles $\theta \in [0, 2\pi]$.

One-shot generation achieves the lowest mean Chamfer Distance of 0.121 ($\pm$0.001 across runs), while graph-level retry shows 0.145 ($\pm$0.001) and node-level retry exhibits 0.188 ($\pm$0.000). However, these metrics are computed over substantially different sample populations. One-shot generation evaluates only the 29.4\% of test cases that naturally produce valid mechanisms (n $\approx$ 25,500), while retry methods successfully generate valid mechanisms for nearly the entire test set (n $\approx$ 84,000 to 86,000).

Several factors contribute to the observed differences in Chamfer Distance. First, the sample composition is fundamentally different. The one-shot generation subset likely represents conditioning curves that are better represented in the training data distribution. These are cases where the network has learned strong associations between curve geometries and corresponding mechanism parameters. These easier cases naturally produce valid mechanisms without retry and also achieve better curve reconstruction. In contrast, retry methods must handle the complete spectrum of test cases, including conditioning curves that may be underrepresented in training data or correspond to more challenging regions of the design space.

Second, Figure~\ref{fig:chamfer}(b) illustrates two important characteristics of the curve comparison that affect Chamfer Distance measurements. The point visualization reveals that while we sample 200 uniformly spaced motor angles, the resulting spatial distribution of end-effector points is not uniform along the curve's arc length. The mechanism's kinematics determine the relationship between motor angle and end-effector velocity. Regions where the mechanism moves quickly (sparse points) and slowly (dense point clusters) create non-uniform sampling that varies with mechanism topology. Different mechanisms generated by retry methods may exhibit different velocity profiles, affecting spatial point distribution and consequently the Chamfer Distance metric.

Third, and critically, the examples in Figure~\ref{fig:chamfer}(b) demonstrate that generated curves can exhibit rotational and reflectional variations relative to conditioning curves even after normalization. While curves are normalized for center position and scale, they retain rotational orientation and may be reflected. Chamfer Distance is not invariant to rotation or reflection, meaning two curves with identical shapes but different orientations will yield non-zero distances. This is particularly relevant for retry methods, which may generate topologically distinct mechanisms that produce geometrically similar curves under different rotations or reflections. For instance, a mechanism synthesized through multiple retry attempts might trace the same curve shape but rotated by some angle or reflected across an axis, resulting in higher Chamfer Distance despite achieving the desired geometric form.

Despite these factors, the absolute differences in Chamfer Distance between strategies remain relatively modest. Node-level retry achieves a mean distance of 0.188 while successfully generating valid mechanisms for 99.6\% of test cases. When considering that this metric is sensitive to point sampling density, rotation, and reflection, a distance of 0.188 demonstrates that the model maintains reasonable curve fidelity even for challenging cases. The consistent pattern across three independent runs (maximum standard deviation of 0.001) confirms this is a systematic characteristic rather than random variation.

These results demonstrate that retry mechanisms enable the model to successfully handle the full diversity of test cases, producing mechanically valid linkages that capture the essential geometric characteristics of conditioning curves, albeit sometimes with rotational or reflectional transformations. While Chamfer Distance increases when including challenging cases and accounting for orientation variations, the ability to generate mechanically valid linkages for nearly all conditioning curves represents a substantial practical advantage over one-shot generation, which succeeds only on a limited subset of easier cases. For practical mechanism design applications, generating a valid mechanism with the correct curve shape under a different orientation is often acceptable, as the mechanism can be physically rotated or reflected during installation.

\section{Discussion}
\subsection{Generation Diversity}
\begin{figure}[!h]
    \includegraphics[width=\textwidth]{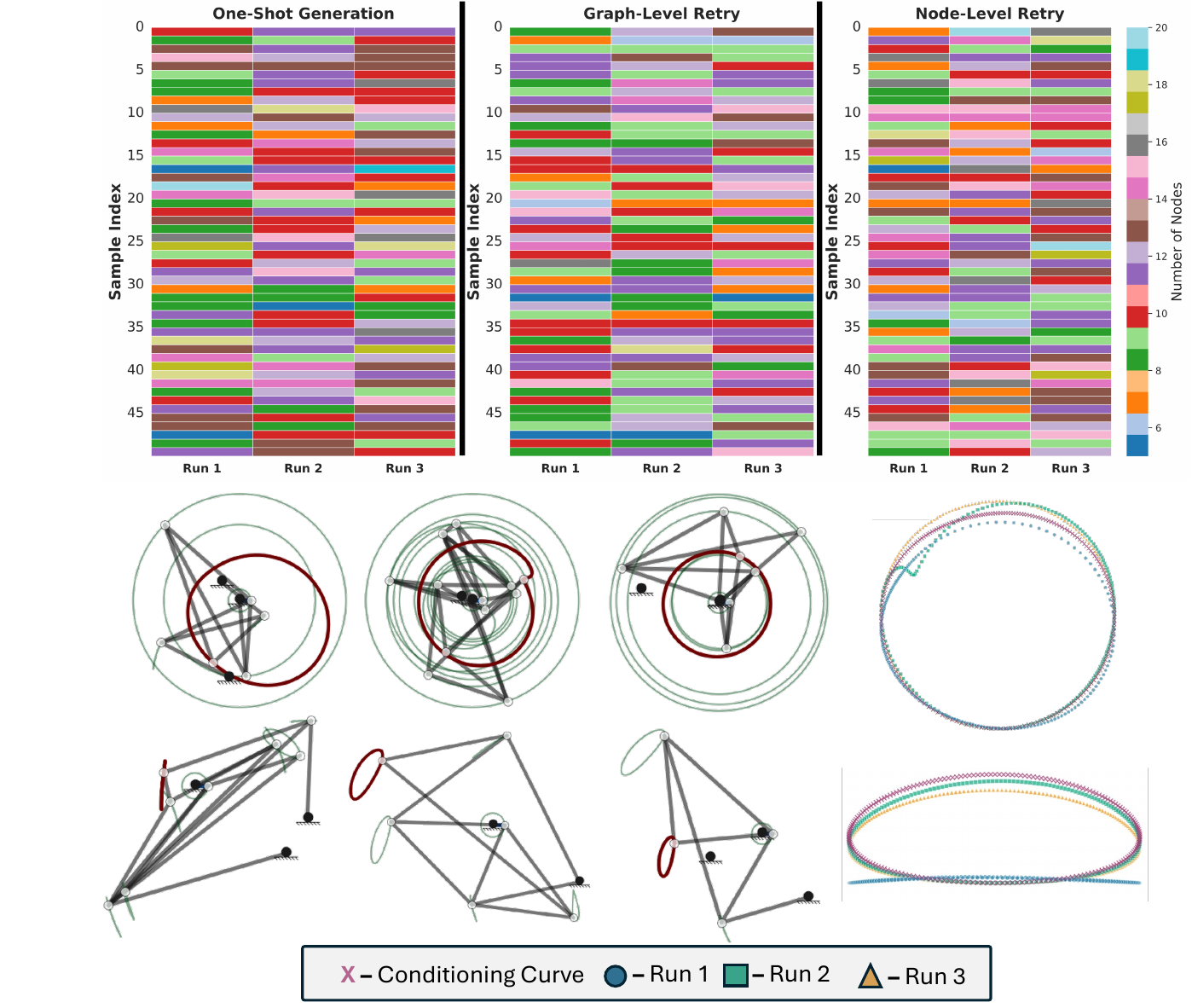}
    
    \caption{\textbf{Mechanism topology diversity across retry strategies.} Top: Heatmaps showing the number of nodes generated for the same conditioning curves (sample indices) across three independent runs. Each row represents a single conditioning curve, with columns showing the node count from each run. Color intensity indicates mechanism complexity (number of nodes). Variation in colors across runs for the same sample demonstrates that the stochastic generation process produces mechanisms with different topological structures for identical conditioning curves. Bottom: Representative examples showing mechanisms with different topologies (indicated by varying node counts and link configurations) that trace similar curves. Left three panels show complete mechanism structures for different samples, while the rightmost panel shows curve overlays where the conditioning curve (X markers) is traced by mechanisms from three runs (circles, squares, triangles) with different topologies but similar geometric outputs.}

    \label{fig:diversity}
\end{figure}

The stochastic nature of autoregressive-diffusion based generation enables our model to produce diverse mechanism topologies for the same conditioning curve, reflecting the one-to-many nature of the inverse design problem. Figure~\ref{fig:diversity} demonstrates this diversity through both quantitative and qualitative analyses.

The heatmaps in the top panel reveal substantial variation in the number of nodes generated across three independent runs for identical conditioning curves. Each row represents a single test sample (conditioning curve), and the color variation across columns (runs) indicates that different mechanism topologies are synthesized for the same design target. For instance, in one-shot generation, sample index 10 produces mechanisms with varying node counts across the three runs, as evidenced by different colors in that row. This pattern is consistent across all three generation strategies, though the extent of variation differs.

Notably, graph-level and node-level retry strategies exhibit greater topological diversity compared to one-shot generation. This increased diversity arises from the retry mechanism's exploration of the learned distribution: when a generated graph fails validation, the stochastic resampling process may select a topologically distinct mechanism that still satisfies the conditioning curve constraint. The heatmaps show that retry methods generate mechanisms spanning a wider range of node counts (4 to 20 nodes) compared to one-shot generation, which tends to produce more consistent complexity levels for the same conditioning curve.

The bottom panel provides visual confirmation of this diversity through representative mechanism examples. The leftmost three panels show complete mechanism structures for different samples, where the same conditioning curve (shown in maroon) is traced by mechanisms with markedly different topologies. These mechanisms vary in the number of links, joint configurations, and overall structural complexity, yet all produce curves that closely approximate the conditioning target. The rightmost panel directly visualizes this diversity by overlaying curves traced by mechanisms from three independent runs (shown as circles, squares, and triangles) alongside the conditioning curve (X markers). Despite originating from different mechanism topologies with varying node counts, the generated curves align closely with each other and the conditioning target, demonstrating that the model successfully learns multiple valid solutions to the inverse design problem.

This topological diversity is practically valuable for mechanism design applications. Different topologies offer distinct advantages in terms of manufacturability, kinematic performance, workspace constraints, and structural robustness. By generating multiple valid solutions for a single design specification, our approach provides engineers with a portfolio of candidate designs rather than a single solution, enabling selection based on secondary criteria such as compactness, link length ratios, or joint placement constraints. The consistent generation of diverse yet valid mechanisms across independent runs demonstrates that this diversity is a reliable characteristic of the learned generative model rather than random variation.

Furthermore, the diversity analysis validates our earlier interpretation of the Chamfer Distance results. The ability to generate topologically distinct mechanisms for the same curve explains why retry methods show higher average Chamfer Distance: the stochastic sampling process explores the space of valid mechanisms, potentially selecting solutions that deviate slightly from the conditioning curve but offer different topological characteristics. This exploration-exploitation trade-off is inherent to the one-to-many inverse problem and represents a feature rather than a limitation of our approach.

\subsection{Curve Embedding Space Analysis}
\begin{figure}[!h]
    \includegraphics[width=\textwidth]{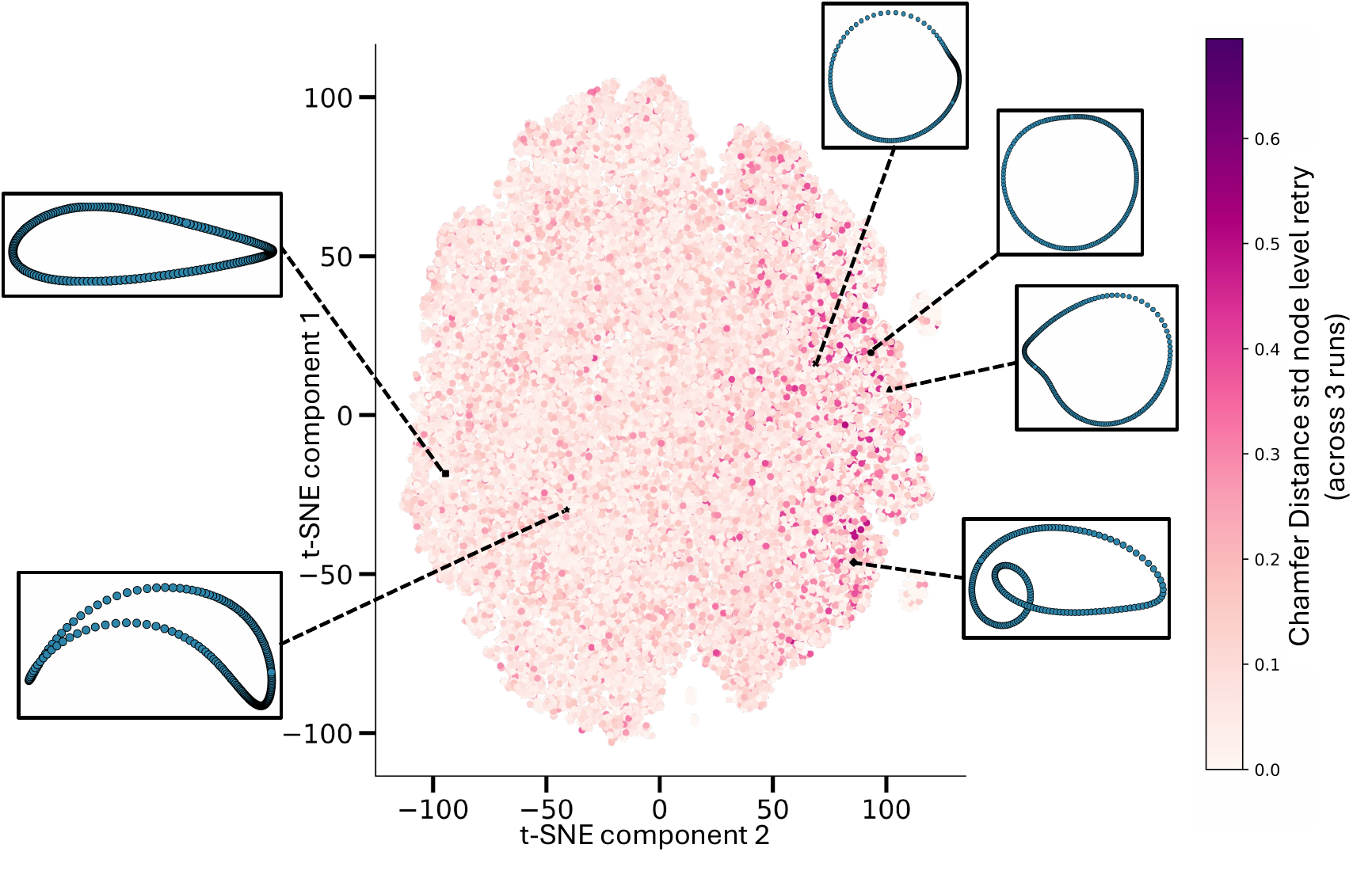}
    
    \caption{\textbf{t-SNE visualization of curve embedding space colored by generation consistency.} Each point represents a conditioning curve from the test set, encoded using the PointBert-style curve encoder and projected to 2D via t-SNE. Color indicates the standard deviation of Chamfer Distance across three independent runs for node-level retry, where darker purple shows higher variance (inconsistent reproduction) and lighter pink shows lower variance (consistent reproduction). Inset examples show representative curves from different regions of the embedding space. The continuous distribution of curves without distinct cluster boundaries correlates with higher variance, particularly in regions where geometrically similar curves are tightly packed. This suggests that well-separated clusters would provide clearer conditioning signals for the autoregressive diffusion process.}

    \label{fig:tsne}
\end{figure}
To understand which types of curves exhibit higher variability in generation quality, we analyze the learned curve embedding space using t-SNE dimensionality reduction. We encode all conditioning curves from the test set using our PointBert-style curve encoder and visualize the resulting high-dimensional embeddings in 2D space (Figure~\ref{fig:tsne}). Each point represents a conditioning curve, and points that are close together in the embedding space correspond to geometrically similar curves. The visualization is colored by the standard deviation of Chamfer Distance across the three independent runs for node-level retry, where darker purple regions indicate curves with high variance in reconstruction accuracy and lighter pink regions show curves that were consistently reproduced.

The embedding space naturally clusters curves by geometric similarity, as shown by the inset examples. Different regions of the t-SNE space correspond to different curve archetypes such as circular paths, elongated ellipses, and more complex multi-lobed shapes. This demonstrates that the PointBert encoder successfully learns meaningful geometric representations that group similar curve shapes together.

The most striking observation is the lack of distinct, well-separated clusters. Instead, conditioning curves are distributed in a continuous embedding space where similar geometries blend smoothly into one another without clear boundaries. This continuous distribution has important implications for the autoregressive diffusion generation process. High-variance regions (darker purple) predominantly occur in areas where curves have very minor geometric differences and exist in this continuous gradient. When curves differ only subtly in their geometric features, such as slight variations in ellipse eccentricity, minor changes in loop size, or small shifts in curve orientation, the stochastic generation process produces inconsistent reconstruction quality across runs.

When the embedding space lacks clear cluster separation, the autoregressive diffusion model receives ambiguous conditioning signals. During node-by-node generation, the model must decide at each step which mechanism topology to construct based on the conditioning vector. However, when curves occupy a continuous manifold rather than discrete clusters, small perturbations in the stochastic diffusion process can lead the model toward different design trajectories, resulting in inconsistent outputs across runs. The model struggles to make fine distinctions between nearly identical curves in the continuous space, leading to higher variance in which mechanism topology is selected and how accurately the curve is reproduced.

Conversely, regions with lower variance (lighter pink) tend to correspond to areas where conditioning curves are more geometrically distinct from their neighbors. In these regions, the curves have clearer geometric differences that the model has learned to distinguish more reliably, leading to more consistent generation outcomes. However, even these regions exist within the overall continuous distribution rather than forming truly separated clusters.

A key insight for model improvement emerges from this observation. If the curve embedding space could be restructured to have distinct, well-separated clusters rather than a continuous distribution, the autoregressive diffusion process would receive much clearer guidance about where to proceed during generation. Each cluster would represent a distinct curve archetype with clear boundaries, providing the diffusion model with unambiguous waypoints for mechanism synthesis. This would reduce sensitivity to stochastic noise and improve consistency, as the model would know definitively which design family it should be targeting at each generation step.

The continuous nature of the current embedding space means that nearly identical curves with only minor differences have nearly identical embeddings, yet may require different mechanism solutions or lead to different stochastic outcomes. In contrast, a clustered embedding space would group curves with similar design solutions together while maintaining clear separations between fundamentally different curve families. This would give the diffusion model discrete, unambiguous targets rather than points in a smooth manifold, reducing the ambiguity that leads to generation variance.

This analysis suggests several potential improvements. Vector quantization of the curve embeddings could create discrete latent codes that enforce natural boundaries. Contrastive learning approaches could push dissimilar curves apart while pulling similar ones together, creating explicit cluster boundaries. Prototype-based encoding methods could map curves to a discrete set of learned prototypes rather than continuous embeddings. Cluster-aware training that explicitly encourages the encoder to produce well-separated, distinct clusters could also address this issue. Such modifications would provide the autoregressive diffusion model with clearer conditioning signals, reducing ambiguity about which design path to follow and improving generation consistency across runs.

\subsection{Causal Transformers as graph networks}
Graph Neural Networks (GNNs) rely on message passing, an iterative process in which each node aggregates information from its neighbors to update its state. This mechanism allows dynamic contextualization: when a new node or edge is added to the graph, message passing propagates information through the structure and updates the existing node embeddings. Each node representation evolves as the graph grows, reflecting both local and global dependencies. In static graphs, this process converges to an equilibrium representation, while in dynamic or incrementally growing graphs, message passing continues to refine the latent state as new information is introduced.

Sequential graph generation introduces a key constraint that does not exist in typical GNNs. Once a node or edge is generated, its state cannot be modified afterward. The model must commit to a representation of the existing nodes before the full graph structure is available. This breaks the symmetry of message passing. Instead of performing iterative updates across all nodes, information can only flow forward in the order of generation.

This constraint creates a causal structure in the graph. The generation of a node $v_t$ depends on the embeddings of previously generated nodes $v_{<t}$, but those earlier nodes cannot be updated based on $v_t$. This reflects a tension between generative causality (the temporal sequence of generation) and structural relationality (the dependencies implied by graph connectivity). Sequential models therefore trade off dynamic updates for autoregressive consistency.

Transformers can be interpreted as operating on a complete directed graph where each token or node attends to every other node through learned edge weights that are determined by attention scores. The attention mechanism functions in a way that resembles message passing: query-key interactions define edge importance, and value projections determine the content of the messages exchanged. Unlike classical GNNs where edges are sparse and predefined, Transformers compute soft, learned connectivity, effectively building a fully connected graph whose edge weights depend on context.

In autoregressive Transformers, such as those used for language or sequential graph generation, the attention graph is causally masked. Each node can only receive messages from previously generated nodes, reflecting the sequential generation constraint. Once a node has been produced, it cannot be updated by future information, which parallels the limitation seen in sequential graph construction.

Viewing Transformers as GNNs reveals both their strengths and their limitations for graph generation. Their dense attention mechanism allows flexible, content-dependent relational modeling, which can be interpreted as dynamic, learnable message passing. However, causal masking enforces a one-directional flow of information. This makes Transformers well suited for tasks that require ordered generation, such as molecular graph construction or program synthesis, but less suitable for problems that need iterative refinement or bidirectional dependency propagation.

Hybrid architectures, including Graph-Transformer decoders and diffusion-based graph models, attempt to overcome this limitation by alternating between autoregressive generation and retroactive message passing. This approach imitates the bidirectional update behavior of traditional GNNs.

---------------------------------------------------------------------------------------






\bibliography{achemso-demo}

\providecommand{\latin}[1]{#1}
\makeatletter
\providecommand{\doi}
  {\begingroup\let\do\@makeother\dospecials
  \catcode`\{=1 \catcode`\}=2 \doi@aux}
\providecommand{\doi@aux}[1]{\endgroup\texttt{#1}}
\makeatother
\providecommand*\mcitethebibliography{\thebibliography}
\csname @ifundefined\endcsname{endmcitethebibliography}  {\let\endmcitethebibliography\endthebibliography}{}
\begin{mcitethebibliography}{58}
\providecommand*\natexlab[1]{#1}
\providecommand*\mciteSetBstSublistMode[1]{}
\providecommand*\mciteSetBstMaxWidthForm[2]{}
\providecommand*\mciteBstWouldAddEndPuncttrue
  {\def\EndOfBibitem{\unskip.}}
\providecommand*\mciteBstWouldAddEndPunctfalse
  {\let\EndOfBibitem\relax}
\providecommand*\mciteSetBstMidEndSepPunct[3]{}
\providecommand*\mciteSetBstSublistLabelBeginEnd[3]{}
\providecommand*\EndOfBibitem{}
\mciteSetBstSublistMode{f}
\mciteSetBstMaxWidthForm{subitem}{(\alph{mcitesubitemcount})}
\mciteSetBstSublistLabelBeginEnd
  {\mcitemaxwidthsubitemform\space}
  {\relax}
  {\relax}

\bibitem[Reuleaux(2013)]{reuleaux2013kinematics}
Reuleaux,~F. \emph{The kinematics of machinery: outlines of a theory of machines}; Courier Corporation, 2013\relax
\mciteBstWouldAddEndPuncttrue
\mciteSetBstMidEndSepPunct{\mcitedefaultmidpunct}
{\mcitedefaultendpunct}{\mcitedefaultseppunct}\relax
\EndOfBibitem
\bibitem[Xin and Liu(2014)Xin, and Liu]{xin2014control}
Xin,~X.; Liu,~Y. \emph{Control design and analysis for underactuated robotic systems}; Springer Science \& Business Media, 2014\relax
\mciteBstWouldAddEndPuncttrue
\mciteSetBstMidEndSepPunct{\mcitedefaultmidpunct}
{\mcitedefaultendpunct}{\mcitedefaultseppunct}\relax
\EndOfBibitem
\bibitem[Gallardo-Alvarado and Gallardo-Razo(2022)Gallardo-Alvarado, and Gallardo-Razo]{gallardo2022mechanisms}
Gallardo-Alvarado,~J.; Gallardo-Razo,~J. \emph{Mechanisms: kinematic analysis and applications in robotics}; Academic Press, 2022\relax
\mciteBstWouldAddEndPuncttrue
\mciteSetBstMidEndSepPunct{\mcitedefaultmidpunct}
{\mcitedefaultendpunct}{\mcitedefaultseppunct}\relax
\EndOfBibitem
\bibitem[Benhabib and Dai(1991)Benhabib, and Dai]{benhabib1991mechanical}
Benhabib,~B.; Dai,~M. Mechanical design of a modular robot for industrial applications. \emph{Journal of Manufacturing Systems} \textbf{1991}, \emph{10}, 297--306\relax
\mciteBstWouldAddEndPuncttrue
\mciteSetBstMidEndSepPunct{\mcitedefaultmidpunct}
{\mcitedefaultendpunct}{\mcitedefaultseppunct}\relax
\EndOfBibitem
\bibitem[Kozuka \latin{et~al.}(2013)Kozuka, Arata, Okuda, Onaga, Ohno, Sano, and Fujimoto]{kozuka2013compliant}
Kozuka,~H.; Arata,~J.; Okuda,~K.; Onaga,~A.; Ohno,~M.; Sano,~A.; Fujimoto,~H. A compliant-parallel mechanism with bio-inspired compliant joints for high precision assembly robot. \emph{Procedia Cirp} \textbf{2013}, \emph{5}, 175--178\relax
\mciteBstWouldAddEndPuncttrue
\mciteSetBstMidEndSepPunct{\mcitedefaultmidpunct}
{\mcitedefaultendpunct}{\mcitedefaultseppunct}\relax
\EndOfBibitem
\bibitem[Zhu \latin{et~al.}(2015)Zhu, Huang, and Li]{zhu2015novel}
Zhu,~Y.; Huang,~X.; Li,~S. A novel six degrees-of-freedom parallel manipulator for aircraft fuselage assemble and its trajectory planning. \emph{Journal of the Chinese Institute of Engineers} \textbf{2015}, \emph{38}, 928--937\relax
\mciteBstWouldAddEndPuncttrue
\mciteSetBstMidEndSepPunct{\mcitedefaultmidpunct}
{\mcitedefaultendpunct}{\mcitedefaultseppunct}\relax
\EndOfBibitem
\bibitem[Wu \latin{et~al.}(2017)Wu, Gao, Zhang, and Wang]{wu2017workspace}
Wu,~J.; Gao,~Y.; Zhang,~B.; Wang,~L. Workspace and dynamic performance evaluation of the parallel manipulators in a spray-painting equipment. \emph{Robotics and Computer-Integrated Manufacturing} \textbf{2017}, \emph{44}, 199--207\relax
\mciteBstWouldAddEndPuncttrue
\mciteSetBstMidEndSepPunct{\mcitedefaultmidpunct}
{\mcitedefaultendpunct}{\mcitedefaultseppunct}\relax
\EndOfBibitem
\bibitem[Ma \latin{et~al.}(2022)Ma, Li, Ma, Wang, Shi, Zheng, Cui, Li, Liu, Guo, \latin{et~al.} others]{ma2022recent}
Ma,~X.; Li,~T.; Ma,~J.; Wang,~Z.; Shi,~C.; Zheng,~S.; Cui,~Q.; Li,~X.; Liu,~F.; Guo,~H.; others Recent advances in space-deployable structures in China. \emph{Engineering} \textbf{2022}, \emph{17}, 207--219\relax
\mciteBstWouldAddEndPuncttrue
\mciteSetBstMidEndSepPunct{\mcitedefaultmidpunct}
{\mcitedefaultendpunct}{\mcitedefaultseppunct}\relax
\EndOfBibitem
\bibitem[Lovasz \latin{et~al.}(2014)Lovasz, Ciupe, Modler, Gruescu, Hanke, Maniu, and M{\u{a}}rgineanu]{lovasz2014experimental}
Lovasz,~E.-C.; Ciupe,~V.; Modler,~K.-H.; Gruescu,~C.; Hanke,~U.; Maniu,~I.; M{\u{a}}rgineanu,~D. Experimental design and control approach of an active knee prosthesis with geared linkage. New Advances in Mechanisms, Transmissions and Applications: Proceedings of the Second Conference MeTrApp 2013. 2014; pp 149--156\relax
\mciteBstWouldAddEndPuncttrue
\mciteSetBstMidEndSepPunct{\mcitedefaultmidpunct}
{\mcitedefaultendpunct}{\mcitedefaultseppunct}\relax
\EndOfBibitem
\bibitem[Tran \latin{et~al.}(2022)Tran, Gabert, Hood, and Lenzi]{tran2022lightweight}
Tran,~M.; Gabert,~L.; Hood,~S.; Lenzi,~T. A lightweight robotic leg prosthesis replicating the biomechanics of the knee, ankle, and toe joint. \emph{Science robotics} \textbf{2022}, \emph{7}, eabo3996\relax
\mciteBstWouldAddEndPuncttrue
\mciteSetBstMidEndSepPunct{\mcitedefaultmidpunct}
{\mcitedefaultendpunct}{\mcitedefaultseppunct}\relax
\EndOfBibitem
\bibitem[Muller(1996)]{muller1996novel}
Muller,~M. A novel classification of planar four-bar linkages and its application to the mechanical analysis of animal systems. \emph{Philosophical Transactions of the Royal Society of London. Series B: Biological Sciences} \textbf{1996}, \emph{351}, 689--720\relax
\mciteBstWouldAddEndPuncttrue
\mciteSetBstMidEndSepPunct{\mcitedefaultmidpunct}
{\mcitedefaultendpunct}{\mcitedefaultseppunct}\relax
\EndOfBibitem
\bibitem[Zhao \latin{et~al.}(2016)Zhao, Zhao, and Yan]{zhao2016planar}
Zhao,~D.-J.; Zhao,~J.-S.; Yan,~Z.-F. Planar deployable linkage and its application in overconstrained lift mechanism. \emph{Journal of Mechanisms and Robotics} \textbf{2016}, \emph{8}\relax
\mciteBstWouldAddEndPuncttrue
\mciteSetBstMidEndSepPunct{\mcitedefaultmidpunct}
{\mcitedefaultendpunct}{\mcitedefaultseppunct}\relax
\EndOfBibitem
\bibitem[Phocas \latin{et~al.}(2020)Phocas, Christoforou, and Dimitriou]{phocas2020kinematics}
Phocas,~M.~C.; Christoforou,~E.~G.; Dimitriou,~P. Kinematics and control approach for deployable and reconfigurable rigid bar linkage structures. \emph{Engineering Structures} \textbf{2020}, \emph{208}, 110310\relax
\mciteBstWouldAddEndPuncttrue
\mciteSetBstMidEndSepPunct{\mcitedefaultmidpunct}
{\mcitedefaultendpunct}{\mcitedefaultseppunct}\relax
\EndOfBibitem
\bibitem[Lipson(2008)]{lipson2008evolutionary}
Lipson,~H. Evolutionary synthesis of kinematic mechanisms. \emph{AI EDAM} \textbf{2008}, \emph{22}, 195--205\relax
\mciteBstWouldAddEndPuncttrue
\mciteSetBstMidEndSepPunct{\mcitedefaultmidpunct}
{\mcitedefaultendpunct}{\mcitedefaultseppunct}\relax
\EndOfBibitem
\bibitem[Ramezani \latin{et~al.}(2016)Ramezani, Shi, Chung, and Hutchinson]{ramezani2016bat}
Ramezani,~A.; Shi,~X.; Chung,~S.-J.; Hutchinson,~S. Bat Bot (B2), a biologically inspired flying machine. 2016 IEEE International Conference on Robotics and Automation (ICRA). 2016; pp 3219--3226\relax
\mciteBstWouldAddEndPuncttrue
\mciteSetBstMidEndSepPunct{\mcitedefaultmidpunct}
{\mcitedefaultendpunct}{\mcitedefaultseppunct}\relax
\EndOfBibitem
\bibitem[Plecnik \latin{et~al.}(2017)Plecnik, Haldane, Yim, and Fearing]{plecnik2017design}
Plecnik,~M.~M.; Haldane,~D.~W.; Yim,~J.~K.; Fearing,~R.~S. Design exploration and kinematic tuning of a power modulating jumping monopod. \emph{Journal of Mechanisms and Robotics} \textbf{2017}, \emph{9}, 011009\relax
\mciteBstWouldAddEndPuncttrue
\mciteSetBstMidEndSepPunct{\mcitedefaultmidpunct}
{\mcitedefaultendpunct}{\mcitedefaultseppunct}\relax
\EndOfBibitem
\bibitem[Plecnik and Michael~McCarthy(2014)Plecnik, and Michael~McCarthy]{plecnik2014numerical}
Plecnik,~M.~M.; Michael~McCarthy,~J. Numerical synthesis of six-bar linkages for mechanical computation. \emph{Journal of Mechanisms and Robotics} \textbf{2014}, \emph{6}, 031012\relax
\mciteBstWouldAddEndPuncttrue
\mciteSetBstMidEndSepPunct{\mcitedefaultmidpunct}
{\mcitedefaultendpunct}{\mcitedefaultseppunct}\relax
\EndOfBibitem
\bibitem[Kim and Kim(2014)Kim, and Kim]{kim2014topology}
Kim,~S.~I.; Kim,~Y.~Y. Topology optimization of planar linkage mechanisms. \emph{International Journal for Numerical Methods in Engineering} \textbf{2014}, \emph{98}, 265--286\relax
\mciteBstWouldAddEndPuncttrue
\mciteSetBstMidEndSepPunct{\mcitedefaultmidpunct}
{\mcitedefaultendpunct}{\mcitedefaultseppunct}\relax
\EndOfBibitem
\bibitem[Tuttle(1996)]{tuttle1996generation}
Tuttle,~E. Generation of planar kinematic chains. \emph{Mechanism and Machine Theory} \textbf{1996}, \emph{31}, 729--748\relax
\mciteBstWouldAddEndPuncttrue
\mciteSetBstMidEndSepPunct{\mcitedefaultmidpunct}
{\mcitedefaultendpunct}{\mcitedefaultseppunct}\relax
\EndOfBibitem
\bibitem[Vasiliu and Yannou(2001)Vasiliu, and Yannou]{vasiliu2001dimensional}
Vasiliu,~A.; Yannou,~B. Dimensional synthesis of planar mechanisms using neural networks: application to path generator linkages. \emph{Mechanism and Machine Theory} \textbf{2001}, \emph{36}, 299--310\relax
\mciteBstWouldAddEndPuncttrue
\mciteSetBstMidEndSepPunct{\mcitedefaultmidpunct}
{\mcitedefaultendpunct}{\mcitedefaultseppunct}\relax
\EndOfBibitem
\bibitem[Lee \latin{et~al.}(2024)Lee, Kim, and Kang]{lee2024deep}
Lee,~S.; Kim,~J.; Kang,~N. Deep generative model-based synthesis framework of four-bar linkage mechanisms with target conditions. \emph{Journal of Computational Design and Engineering} \textbf{2024}, \emph{11}, 318--332\relax
\mciteBstWouldAddEndPuncttrue
\mciteSetBstMidEndSepPunct{\mcitedefaultmidpunct}
{\mcitedefaultendpunct}{\mcitedefaultseppunct}\relax
\EndOfBibitem
\bibitem[Balli and Chand(2002)Balli, and Chand]{balli2002defects}
Balli,~S.~S.; Chand,~S. Defects in link mechanisms and solution rectification. \emph{Mechanism and Machine Theory} \textbf{2002}, \emph{37}, 851--876\relax
\mciteBstWouldAddEndPuncttrue
\mciteSetBstMidEndSepPunct{\mcitedefaultmidpunct}
{\mcitedefaultendpunct}{\mcitedefaultseppunct}\relax
\EndOfBibitem
\bibitem[McCarthy(2000)]{mccarthy2000geometric}
McCarthy,~J.~M. \emph{Geometric design of linkages}; Springer, 2000\relax
\mciteBstWouldAddEndPuncttrue
\mciteSetBstMidEndSepPunct{\mcitedefaultmidpunct}
{\mcitedefaultendpunct}{\mcitedefaultseppunct}\relax
\EndOfBibitem
\bibitem[Wampler \latin{et~al.}(1992)Wampler, Morgan, and Sommese]{wampler1992complete}
Wampler,~C.~W.; Morgan,~A.; Sommese,~A.~J. Complete solution of the nine-point path synthesis problem for four-bar linkages. \textbf{1992}, \relax
\mciteBstWouldAddEndPunctfalse
\mciteSetBstMidEndSepPunct{\mcitedefaultmidpunct}
{}{\mcitedefaultseppunct}\relax
\EndOfBibitem
\bibitem[Pan \latin{et~al.}(2023)Pan, Liu, Gao, and Manocha]{pan2023joint}
Pan,~Z.; Liu,~M.; Gao,~X.; Manocha,~D. Joint search of optimal topology and trajectory for planar linkages. \emph{The International Journal of Robotics Research} \textbf{2023}, \emph{42}, 176--195\relax
\mciteBstWouldAddEndPuncttrue
\mciteSetBstMidEndSepPunct{\mcitedefaultmidpunct}
{\mcitedefaultendpunct}{\mcitedefaultseppunct}\relax
\EndOfBibitem
\bibitem[Fogelson \latin{et~al.}(2023)Fogelson, Tucker, and Cagan]{fogelson2023gcp}
Fogelson,~M.~B.; Tucker,~C.; Cagan,~J. GCP-HOLO: Generating high-order linkage graphs for path synthesis. \emph{Journal of Mechanical Design} \textbf{2023}, \emph{145}, 073303\relax
\mciteBstWouldAddEndPuncttrue
\mciteSetBstMidEndSepPunct{\mcitedefaultmidpunct}
{\mcitedefaultendpunct}{\mcitedefaultseppunct}\relax
\EndOfBibitem
\bibitem[Deshpande and Purwar(2021)Deshpande, and Purwar]{deshpande2021image}
Deshpande,~S.; Purwar,~A. An image-based approach to variational path synthesis of linkages. \emph{Journal of Computing and Information Science in Engineering} \textbf{2021}, \emph{21}, 021005\relax
\mciteBstWouldAddEndPuncttrue
\mciteSetBstMidEndSepPunct{\mcitedefaultmidpunct}
{\mcitedefaultendpunct}{\mcitedefaultseppunct}\relax
\EndOfBibitem
\bibitem[Nobari \latin{et~al.}(2024)Nobari, Srivastava, Gutfreund, Xu, and Ahmed]{nobari2024link}
Nobari,~A.~H.; Srivastava,~A.; Gutfreund,~D.; Xu,~K.; Ahmed,~F. Link: Learning joint representations of design and performance spaces through contrastive learning for mechanism synthesis. \emph{arXiv preprint arXiv:2405.20592} \textbf{2024}, \relax
\mciteBstWouldAddEndPunctfalse
\mciteSetBstMidEndSepPunct{\mcitedefaultmidpunct}
{}{\mcitedefaultseppunct}\relax
\EndOfBibitem
\bibitem[Bolanos \latin{et~al.}(2025)Bolanos, Ataei, and Jayaraman]{bolanos2025mechaformer}
Bolanos,~D.; Ataei,~M.; Jayaraman,~P.~K. MechaFormer: Sequence Learning for Kinematic Mechanism Design Automation. \emph{arXiv preprint arXiv:2508.09005} \textbf{2025}, \relax
\mciteBstWouldAddEndPunctfalse
\mciteSetBstMidEndSepPunct{\mcitedefaultmidpunct}
{}{\mcitedefaultseppunct}\relax
\EndOfBibitem
\bibitem[Deshpande and Purwar(2019)Deshpande, and Purwar]{deshpande2019computational}
Deshpande,~S.; Purwar,~A. Computational creativity via assisted variational synthesis of mechanisms using deep generative models. \emph{Journal of Mechanical Design} \textbf{2019}, \emph{141}, 121402\relax
\mciteBstWouldAddEndPuncttrue
\mciteSetBstMidEndSepPunct{\mcitedefaultmidpunct}
{\mcitedefaultendpunct}{\mcitedefaultseppunct}\relax
\EndOfBibitem
\bibitem[Khan \latin{et~al.}(2015)Khan, Ullah, and Al-Grafi]{khan2015dimensional}
Khan,~N.; Ullah,~I.; Al-Grafi,~M. Dimensional synthesis of mechanical linkages using artificial neural networks and Fourier descriptors. \emph{Mechanical Sciences} \textbf{2015}, \emph{6}, 29--34\relax
\mciteBstWouldAddEndPuncttrue
\mciteSetBstMidEndSepPunct{\mcitedefaultmidpunct}
{\mcitedefaultendpunct}{\mcitedefaultseppunct}\relax
\EndOfBibitem
\bibitem[Goodfellow \latin{et~al.}(2014)Goodfellow, Pouget-Abadie, Mirza, Xu, Warde-Farley, Ozair, Courville, and Bengio]{goodfellow2014generative}
Goodfellow,~I.~J.; Pouget-Abadie,~J.; Mirza,~M.; Xu,~B.; Warde-Farley,~D.; Ozair,~S.; Courville,~A.; Bengio,~Y. Generative adversarial nets. \emph{Advances in neural information processing systems} \textbf{2014}, \emph{27}\relax
\mciteBstWouldAddEndPuncttrue
\mciteSetBstMidEndSepPunct{\mcitedefaultmidpunct}
{\mcitedefaultendpunct}{\mcitedefaultseppunct}\relax
\EndOfBibitem
\bibitem[Nurizada \latin{et~al.}(2025)Nurizada, Lyu, and Purwar]{nurizada2025path}
Nurizada,~A.; Lyu,~Z.; Purwar,~A. Path generative model based on conditional $\beta$-variational auto encoder for four-bar mechanism design. \emph{Journal of Mechanisms and Robotics} \textbf{2025}, \emph{17}, 061004\relax
\mciteBstWouldAddEndPuncttrue
\mciteSetBstMidEndSepPunct{\mcitedefaultmidpunct}
{\mcitedefaultendpunct}{\mcitedefaultseppunct}\relax
\EndOfBibitem
\bibitem[Vermeer \latin{et~al.}(2018)Vermeer, Kuppens, and Herder]{vermeer2018kinematic}
Vermeer,~K.; Kuppens,~R.; Herder,~J. Kinematic synthesis using reinforcement learning. International Design Engineering Technical Conferences and Computers and Information in Engineering Conference. 2018; p V02AT03A009\relax
\mciteBstWouldAddEndPuncttrue
\mciteSetBstMidEndSepPunct{\mcitedefaultmidpunct}
{\mcitedefaultendpunct}{\mcitedefaultseppunct}\relax
\EndOfBibitem
\bibitem[Schulman \latin{et~al.}(2017)Schulman, Wolski, Dhariwal, Radford, and Klimov]{schulman2017proximal}
Schulman,~J.; Wolski,~F.; Dhariwal,~P.; Radford,~A.; Klimov,~O. Proximal policy optimization algorithms. \emph{arXiv preprint arXiv:1707.06347} \textbf{2017}, \relax
\mciteBstWouldAddEndPunctfalse
\mciteSetBstMidEndSepPunct{\mcitedefaultmidpunct}
{}{\mcitedefaultseppunct}\relax
\EndOfBibitem
\bibitem[Whitman \latin{et~al.}(2020)Whitman, Bhirangi, Travers, and Choset]{whitman2020modular}
Whitman,~J.; Bhirangi,~R.; Travers,~M.; Choset,~H. Modular robot design synthesis with deep reinforcement learning. Proceedings of the AAAI conference on artificial intelligence. 2020; pp 10418--10425\relax
\mciteBstWouldAddEndPuncttrue
\mciteSetBstMidEndSepPunct{\mcitedefaultmidpunct}
{\mcitedefaultendpunct}{\mcitedefaultseppunct}\relax
\EndOfBibitem
\bibitem[Hansen and Ostermeier(2001)Hansen, and Ostermeier]{hansen2001completely}
Hansen,~N.; Ostermeier,~A. Completely derandomized self-adaptation in evolution strategies. \emph{Evolutionary computation} \textbf{2001}, \emph{9}, 159--195\relax
\mciteBstWouldAddEndPuncttrue
\mciteSetBstMidEndSepPunct{\mcitedefaultmidpunct}
{\mcitedefaultendpunct}{\mcitedefaultseppunct}\relax
\EndOfBibitem
\bibitem[Wu \latin{et~al.}(2020)Wu, Pan, Chen, Long, Zhang, and Yu]{wu2020comprehensive}
Wu,~Z.; Pan,~S.; Chen,~F.; Long,~G.; Zhang,~C.; Yu,~P.~S. A comprehensive survey on graph neural networks. \emph{IEEE transactions on neural networks and learning systems} \textbf{2020}, \emph{32}, 4--24\relax
\mciteBstWouldAddEndPuncttrue
\mciteSetBstMidEndSepPunct{\mcitedefaultmidpunct}
{\mcitedefaultendpunct}{\mcitedefaultseppunct}\relax
\EndOfBibitem
\bibitem[Corso \latin{et~al.}(2024)Corso, Stark, Jegelka, Jaakkola, and Barzilay]{corso2024graph}
Corso,~G.; Stark,~H.; Jegelka,~S.; Jaakkola,~T.; Barzilay,~R. Graph neural networks. \emph{Nature Reviews Methods Primers} \textbf{2024}, \emph{4}, 17\relax
\mciteBstWouldAddEndPuncttrue
\mciteSetBstMidEndSepPunct{\mcitedefaultmidpunct}
{\mcitedefaultendpunct}{\mcitedefaultseppunct}\relax
\EndOfBibitem
\bibitem[Scarselli \latin{et~al.}(2008)Scarselli, Gori, Tsoi, Hagenbuchner, and Monfardini]{scarselli2008graph}
Scarselli,~F.; Gori,~M.; Tsoi,~A.~C.; Hagenbuchner,~M.; Monfardini,~G. The graph neural network model. \emph{IEEE transactions on neural networks} \textbf{2008}, \emph{20}, 61--80\relax
\mciteBstWouldAddEndPuncttrue
\mciteSetBstMidEndSepPunct{\mcitedefaultmidpunct}
{\mcitedefaultendpunct}{\mcitedefaultseppunct}\relax
\EndOfBibitem
\bibitem[Zhou \latin{et~al.}(2020)Zhou, Cui, Hu, Zhang, Yang, Liu, Wang, Li, and Sun]{zhou2020graph}
Zhou,~J.; Cui,~G.; Hu,~S.; Zhang,~Z.; Yang,~C.; Liu,~Z.; Wang,~L.; Li,~C.; Sun,~M. Graph neural networks: A review of methods and applications. \emph{AI open} \textbf{2020}, \emph{1}, 57--81\relax
\mciteBstWouldAddEndPuncttrue
\mciteSetBstMidEndSepPunct{\mcitedefaultmidpunct}
{\mcitedefaultendpunct}{\mcitedefaultseppunct}\relax
\EndOfBibitem
\bibitem[Liu and Zhou(2022)Liu, and Zhou]{liu2022introduction}
Liu,~Z.; Zhou,~J. \emph{Introduction to graph neural networks}; Springer Nature, 2022\relax
\mciteBstWouldAddEndPuncttrue
\mciteSetBstMidEndSepPunct{\mcitedefaultmidpunct}
{\mcitedefaultendpunct}{\mcitedefaultseppunct}\relax
\EndOfBibitem
\bibitem[Jadhav \latin{et~al.}(2023)Jadhav, Berthel, Hu, Panat, Beuth, and Farimani]{jadhav2023stressd}
Jadhav,~Y.; Berthel,~J.; Hu,~C.; Panat,~R.; Beuth,~J.; Farimani,~A.~B. StressD: 2D Stress estimation using denoising diffusion model. \emph{Computer Methods in Applied Mechanics and Engineering} \textbf{2023}, \emph{416}, 116343\relax
\mciteBstWouldAddEndPuncttrue
\mciteSetBstMidEndSepPunct{\mcitedefaultmidpunct}
{\mcitedefaultendpunct}{\mcitedefaultseppunct}\relax
\EndOfBibitem
\bibitem[Li \latin{et~al.}(2025)Li, Patil, Ogoke, Shu, Zhen, Schneier, Buchanan~Jr, and Farimani]{li2025latent}
Li,~Z.; Patil,~S.; Ogoke,~F.; Shu,~D.; Zhen,~W.; Schneier,~M.; Buchanan~Jr,~J.~R.; Farimani,~A.~B. Latent neural PDE solver: A reduced-order modeling framework for partial differential equations. \emph{Journal of Computational Physics} \textbf{2025}, \emph{524}, 113705\relax
\mciteBstWouldAddEndPuncttrue
\mciteSetBstMidEndSepPunct{\mcitedefaultmidpunct}
{\mcitedefaultendpunct}{\mcitedefaultseppunct}\relax
\EndOfBibitem
\bibitem[Jadhav \latin{et~al.}(2024)Jadhav, Berthel, Hu, Panat, Beuth, and Barati~Farimani]{jadhav2024generative}
Jadhav,~Y.; Berthel,~J.; Hu,~C.; Panat,~R.; Beuth,~J.; Barati~Farimani,~A. Generative lattice units with 3d diffusion for inverse design: Glu3d. \emph{Advanced Functional Materials} \textbf{2024}, \emph{34}, 2404165\relax
\mciteBstWouldAddEndPuncttrue
\mciteSetBstMidEndSepPunct{\mcitedefaultmidpunct}
{\mcitedefaultendpunct}{\mcitedefaultseppunct}\relax
\EndOfBibitem
\bibitem[Zhou \latin{et~al.}(2024)Zhou, Li, Schneier, Buchanan~Jr, and Farimani]{zhou2024text2pde}
Zhou,~A.; Li,~Z.; Schneier,~M.; Buchanan~Jr,~J.~R.; Farimani,~A.~B. Text2pde: Latent diffusion models for accessible physics simulation. \emph{arXiv preprint arXiv:2410.01153} \textbf{2024}, \relax
\mciteBstWouldAddEndPunctfalse
\mciteSetBstMidEndSepPunct{\mcitedefaultmidpunct}
{}{\mcitedefaultseppunct}\relax
\EndOfBibitem
\bibitem[Bartsch \latin{et~al.}(2024)Bartsch, Car, Avra, and Farimani]{bartsch2024sculptdiff}
Bartsch,~A.; Car,~A.; Avra,~C.; Farimani,~A.~B. Sculptdiff: Learning robotic clay sculpting from humans with goal conditioned diffusion policy. 2024 IEEE/RSJ International Conference on Intelligent Robots and Systems (IROS). 2024; pp 7307--7314\relax
\mciteBstWouldAddEndPuncttrue
\mciteSetBstMidEndSepPunct{\mcitedefaultmidpunct}
{\mcitedefaultendpunct}{\mcitedefaultseppunct}\relax
\EndOfBibitem
\bibitem[Graves and Farimani(2024)Graves, and Farimani]{graves2024airfoil}
Graves,~R.; Farimani,~A.~B. Airfoil Diffusion: Denoising Diffusion Model For Conditional Airfoil Generation. \emph{arXiv preprint arXiv:2408.15898} \textbf{2024}, \relax
\mciteBstWouldAddEndPunctfalse
\mciteSetBstMidEndSepPunct{\mcitedefaultmidpunct}
{}{\mcitedefaultseppunct}\relax
\EndOfBibitem
\bibitem[Li \latin{et~al.}(2024)Li, Tian, Li, Deng, and He]{li2024autoregressive}
Li,~T.; Tian,~Y.; Li,~H.; Deng,~M.; He,~K. Autoregressive image generation without vector quantization. \emph{Advances in Neural Information Processing Systems} \textbf{2024}, \emph{37}, 56424--56445\relax
\mciteBstWouldAddEndPuncttrue
\mciteSetBstMidEndSepPunct{\mcitedefaultmidpunct}
{\mcitedefaultendpunct}{\mcitedefaultseppunct}\relax
\EndOfBibitem
\bibitem[Qi \latin{et~al.}(2017)Qi, Su, Mo, and Guibas]{qi2017pointnet}
Qi,~C.~R.; Su,~H.; Mo,~K.; Guibas,~L.~J. Pointnet: Deep learning on point sets for 3d classification and segmentation. Proceedings of the IEEE conference on computer vision and pattern recognition. 2017; pp 652--660\relax
\mciteBstWouldAddEndPuncttrue
\mciteSetBstMidEndSepPunct{\mcitedefaultmidpunct}
{\mcitedefaultendpunct}{\mcitedefaultseppunct}\relax
\EndOfBibitem
\bibitem[Yu \latin{et~al.}(2022)Yu, Tang, Rao, Huang, Zhou, and Lu]{yu2022point}
Yu,~X.; Tang,~L.; Rao,~Y.; Huang,~T.; Zhou,~J.; Lu,~J. Point-bert: Pre-training 3d point cloud transformers with masked point modeling. Proceedings of the IEEE/CVF conference on computer vision and pattern recognition. 2022; pp 19313--19322\relax
\mciteBstWouldAddEndPuncttrue
\mciteSetBstMidEndSepPunct{\mcitedefaultmidpunct}
{\mcitedefaultendpunct}{\mcitedefaultseppunct}\relax
\EndOfBibitem
\bibitem[Yamada and Sugiyama(2023)Yamada, and Sugiyama]{yamada2023molecular}
Yamada,~M.; Sugiyama,~M. Molecular graph generation by decomposition and reassembling. \emph{ACS omega} \textbf{2023}, \emph{8}, 19575--19586\relax
\mciteBstWouldAddEndPuncttrue
\mciteSetBstMidEndSepPunct{\mcitedefaultmidpunct}
{\mcitedefaultendpunct}{\mcitedefaultseppunct}\relax
\EndOfBibitem
\bibitem[Liu \latin{et~al.}(2021)Liu, Yan, Oztekin, and Ji]{liu2021graphebm}
Liu,~M.; Yan,~K.; Oztekin,~B.; Ji,~S. Graphebm: Molecular graph generation with energy-based models. \emph{arXiv preprint arXiv:2102.00546} \textbf{2021}, \relax
\mciteBstWouldAddEndPunctfalse
\mciteSetBstMidEndSepPunct{\mcitedefaultmidpunct}
{}{\mcitedefaultseppunct}\relax
\EndOfBibitem
\bibitem[Dong \latin{et~al.}(2023)Dong, Cao, Zhang, Tao, Chen, and Zhang]{dong2023cktgnn}
Dong,~Z.; Cao,~W.; Zhang,~M.; Tao,~D.; Chen,~Y.; Zhang,~X. CktGNN: Circuit graph neural network for electronic design automation. \emph{arXiv preprint arXiv:2308.16406} \textbf{2023}, \relax
\mciteBstWouldAddEndPunctfalse
\mciteSetBstMidEndSepPunct{\mcitedefaultmidpunct}
{}{\mcitedefaultseppunct}\relax
\EndOfBibitem
\bibitem[Ho \latin{et~al.}(2020)Ho, Jain, and Abbeel]{ho2020denoising}
Ho,~J.; Jain,~A.; Abbeel,~P. Denoising diffusion probabilistic models. \emph{Advances in neural information processing systems} \textbf{2020}, \emph{33}, 6840--6851\relax
\mciteBstWouldAddEndPuncttrue
\mciteSetBstMidEndSepPunct{\mcitedefaultmidpunct}
{\mcitedefaultendpunct}{\mcitedefaultseppunct}\relax
\EndOfBibitem
\bibitem[Song \latin{et~al.}(2020)Song, Meng, and Ermon]{song2020denoising}
Song,~J.; Meng,~C.; Ermon,~S. Denoising diffusion implicit models. \emph{arXiv preprint arXiv:2010.02502} \textbf{2020}, \relax
\mciteBstWouldAddEndPunctfalse
\mciteSetBstMidEndSepPunct{\mcitedefaultmidpunct}
{}{\mcitedefaultseppunct}\relax
\EndOfBibitem
\bibitem[Perez \latin{et~al.}(2018)Perez, Strub, De~Vries, Dumoulin, and Courville]{perez2018film}
Perez,~E.; Strub,~F.; De~Vries,~H.; Dumoulin,~V.; Courville,~A. Film: Visual reasoning with a general conditioning layer. Proceedings of the AAAI conference on artificial intelligence. 2018\relax
\mciteBstWouldAddEndPuncttrue
\mciteSetBstMidEndSepPunct{\mcitedefaultmidpunct}
{\mcitedefaultendpunct}{\mcitedefaultseppunct}\relax
\EndOfBibitem
\end{mcitethebibliography}

\end{document}